\documentclass[10pt,twocolumn,letterpaper]{article}
\pdfoutput=1
\usepackage{iccv}
\usepackage{times}
\usepackage{epsfig}
\usepackage{graphicx}
\usepackage{amsmath}
\usepackage{amssymb}
\usepackage{booktabs}
\usepackage{appendix}
\usepackage{multirow}
\usepackage{bbding}
\usepackage{bm}
\usepackage[table]{xcolor}

\usepackage[pagebackref=true,breaklinks=true,letterpaper=true,colorlinks,bookmarks=false]{hyperref}

\iccvfinalcopy 

\def\iccvPaperID{2480} 

\ificcvfinal\fi

\begin{document}

\title{Visually-Prompted Language Model for Fine-Grained Scene Graph Generation in an Open World}

\author{Qifan Yu$^1$ \quad\quad Juncheng Li$^1$\footnotemark[2] \quad\quad  Yu Wu$^2$ \quad\quad  
Siliang Tang$^1$ \quad\quad Wei Ji$^3$ \quad\quad Yueting Zhuang$^1$\\ $^1$Zhejiang University, $^2$Wuhan University, $^3$National University of Singapore\\ {\tt\small \{yuqifan, junchengli, siliang, yzhuang\}@zju.edu.cn}  \\\tt\small yu.wu-3@student.uts.edu.au, jiwei@nus.edu.sg}

\maketitle

\renewcommand{\thefootnote}{\fnsymbol{footnote}} 
\footnotetext[2]{Corresponding Authors.}
\renewcommand{\thefootnote}{\arabic{footnote}}

\maketitle
\ificcvfinal\thispagestyle{empty}\fi

\begin{abstract}
\textbf{S}cene \textbf{G}raph \textbf{G}eneration~(SGG) aims to extract~\textless subject, predicate, object\textgreater~relationships in images for vision understanding. Although recent works have made steady progress on SGG, they still suffer long-tail distribution issues that tail-predicates are more costly to train and hard to distinguish due to a small amount of annotated data compared to frequent predicates. Existing re-balancing strategies try to handle it via prior rules but are still confined to pre-defined conditions, which are not scalable for various models and datasets. In this paper, we propose a \textbf{C}ross-mod\textbf{a}l predi\textbf{Ca}te b\textbf{o}osting~(\textbf{CaCao}) framework, where a visually-prompted language model is learned to generate diverse fine-grained predicates in a low-resource way. The proposed CaCao can be applied in a plug-and-play fashion and automatically strengthen existing SGG to tackle the long-tailed problem. Based on that, we further introduce a novel \textbf{E}ntangled cross-modal \textbf{p}rompt approach for open-world pred\textbf{i}cate s\textbf{c}ene graph generation~(\textbf{Epic}), where models can generalize to unseen predicates in a zero-shot manner. Comprehensive experiments on three benchmark datasets show that CaCao consistently boosts the performance of multiple scene graph generation models in a model-agnostic way. Moreover, our Epic achieves competitive performance on open-world predicate prediction. The data and code for this paper are publicly available.\footnote{\url{https://github.com/Yuqifan1117/CaCao}}
\end{abstract}

\section{Introduction}
\label{sec:intro}

\begin{figure}[!t]
   \begin{subfigure}{0.32\linewidth}
    \centering
    \includegraphics[width=1.\linewidth]{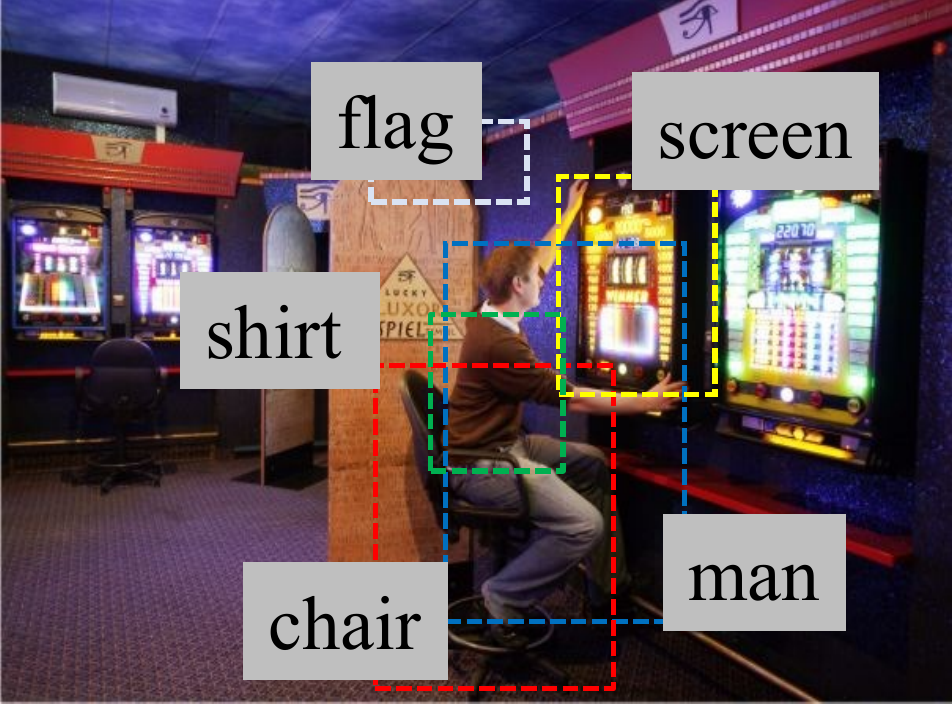}
    \caption{Detected Image}
    \label{input}
  \end{subfigure}
  \begin{subfigure}{0.33\linewidth}
    \centering
    \includegraphics[width=1.0\linewidth]{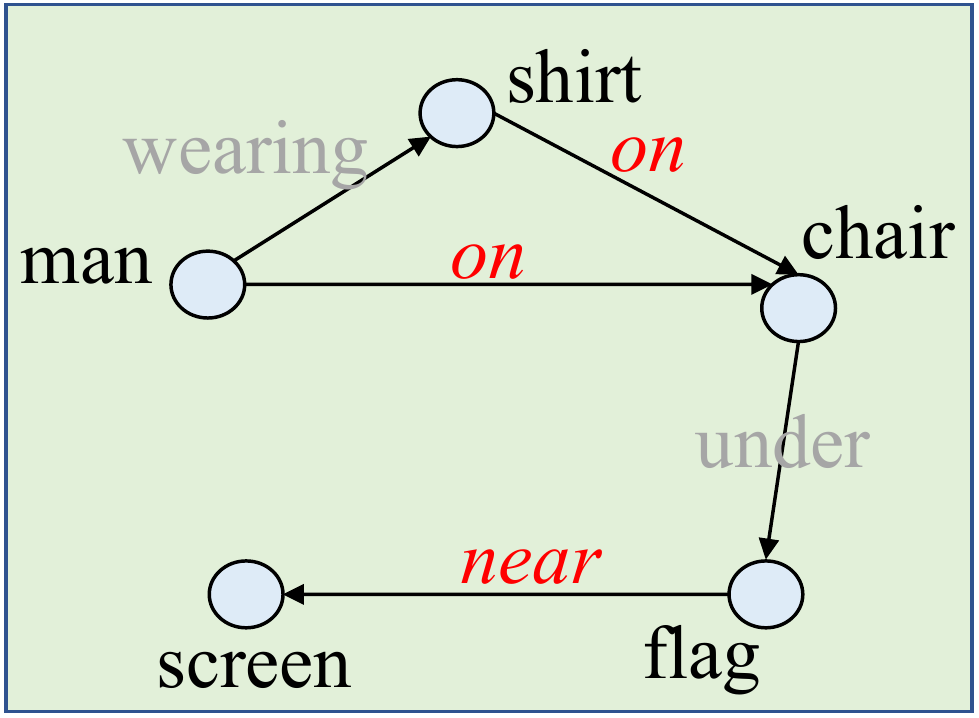}
    \caption{Base SGG}
    \label{sg1}
  \end{subfigure}
    \begin{subfigure}{0.33\linewidth}
    \centering
    \includegraphics[width=1.0\linewidth]{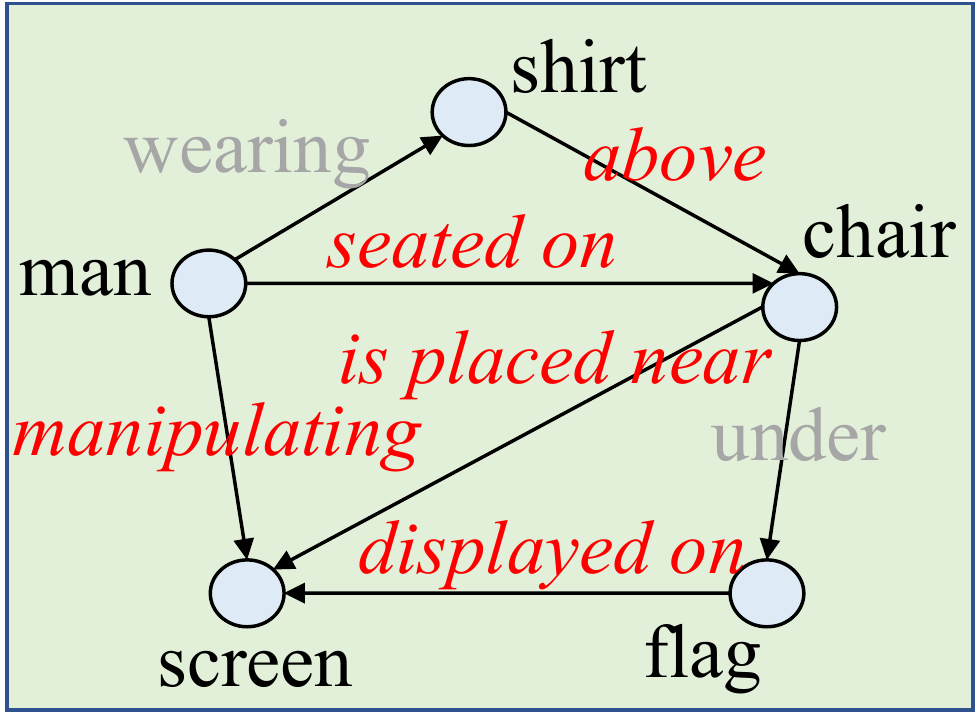}
    \caption{Enhanced SGG}
    \label{sg2}
  \end{subfigure}
  \begin{subfigure}{0.50\linewidth}

   \includegraphics[width=1.0\linewidth]{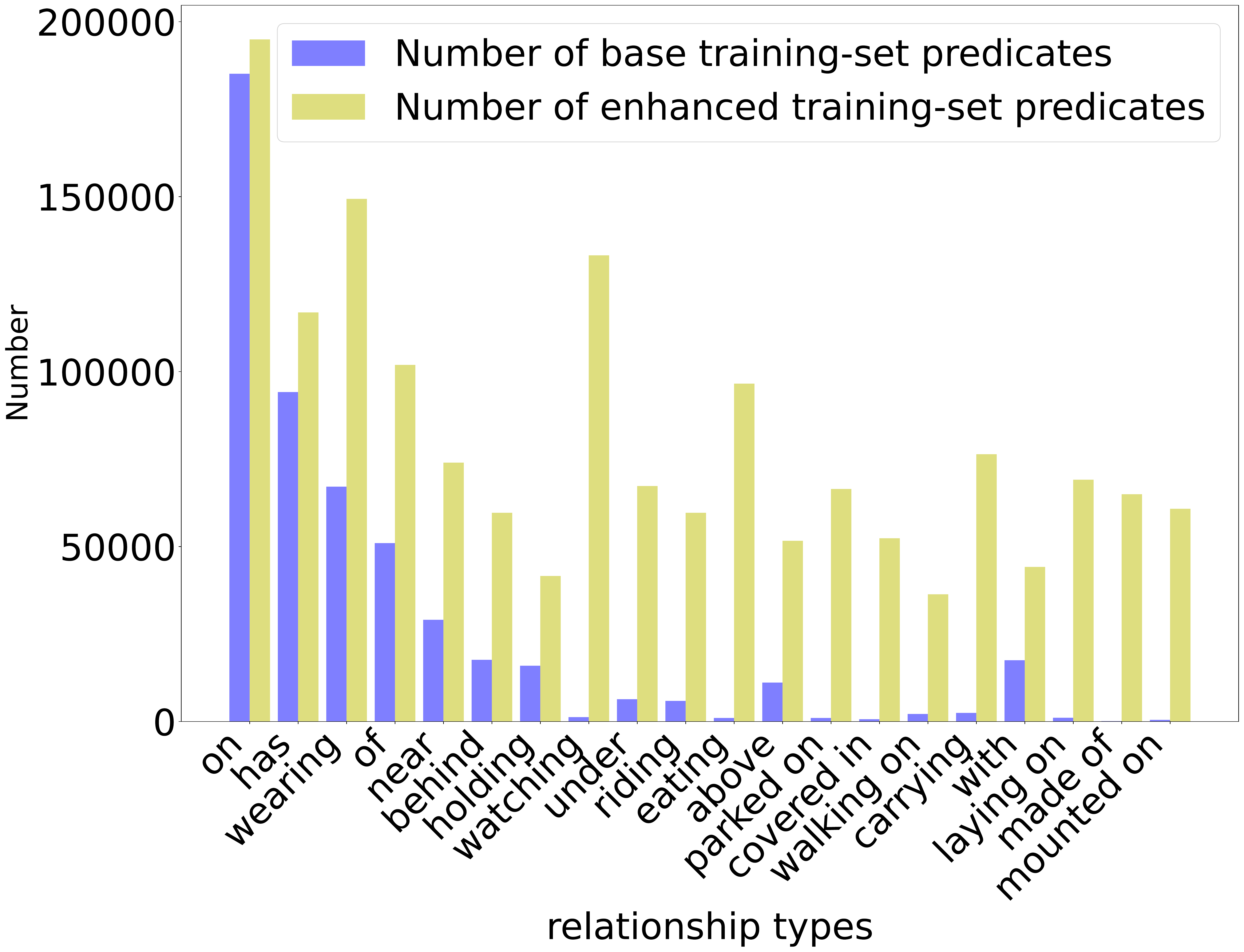}
    \caption{Long-tail predicate distribution}
   \label{long-tail distribution}
   \end{subfigure}
   \begin{subfigure}{0.48\linewidth}

   \includegraphics[width=1.0\linewidth]{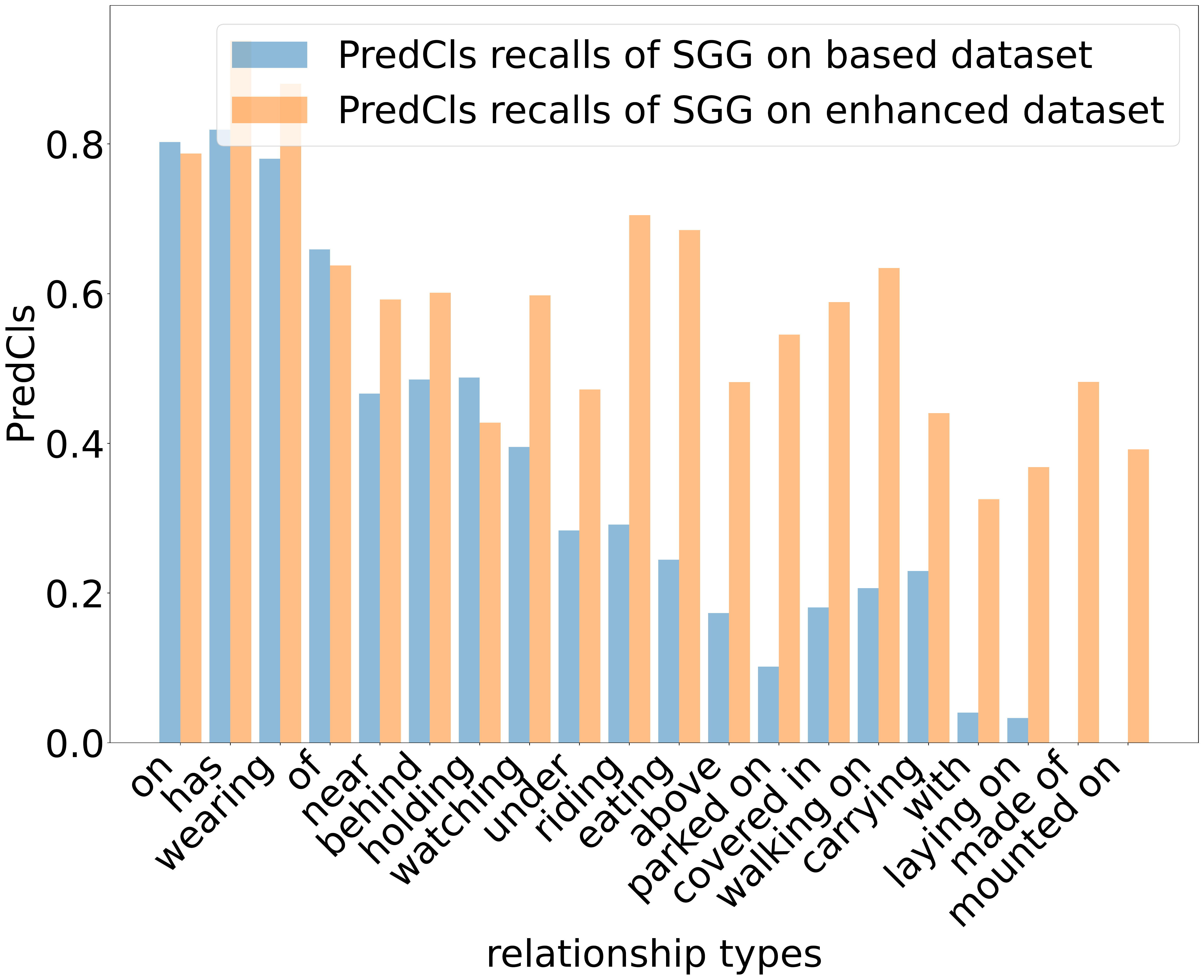}
    \caption{PredCls of different predicates}
   \label{predcls-recall}
   \end{subfigure}
   \vspace{-0.1cm}
    \caption{\textbf{Illustration of handling long-tail distribution problem by cross-modal predicate boosting in Visual Genome.}~(b) and (c) show scene graphs enhanced by visual knowledge generating more informative predicates in long-tail distribution.~(d) indicates the imbalance of predicates due to the long-tailed distribution in the training set.~(e) For prediction of scene graph relationships~(PredCls), our CaCao framework can obtain consistent improvement on both head predicates and tail predicates.} 
    \label{sgg-example}
\end{figure}

Scene graph generation~(SGG) aims to detect visual relationships in real-world images, which consist of the subject, predicate, and object~(\textit{i.e.,}~\textbf{subject}:~\textit{flag},~\textbf{predicate}:~\textit{displayed on},~\textbf{object}:~\textit{screen} in Figure~\ref{sgg-example}~(a)). Since scene graphs bridge the gap between raw pixels and high-level visual semantics, SGG has been widely used in a variety of visual scene analysis and understanding tasks~\cite{chang2021comprehensive, li2021adaptive, li2022compositional, li2023variational}, such as visual question answering~\cite{jiang2020defense, han2021greedy}, image captioning~\cite{zhang2021consensus,yang2022reformer}, and 3D scene understanding~\cite{gothoskar20213dp3, zhang2021exploiting}.

Recently, various methods~\cite{chen2019knowledge,zellers2018neural,zareian2020weakly,tang2019learning,yao2021visual,yan2020pcpl,desai2021learning} have been proposed to improve the SGG performance, but still tend to predict frequent but uninformative predicates due to the long-tailed distribution of predicates in SGG datasets~\cite{xu2017scene,li2021bipartite,yu2020cogtree}. In a way, those approaches degenerate into a trivial solution, which undermines the application of SGG. As shown in Figure~\ref{sgg-example}~(d), in the Visual Genome~\cite{xu2017scene}, the top 20\% of predicate categories account for almost 90\% of samples, while other tail fine-grained predicates lack sufficient training data. Accordingly, the PredCls recalls of SGG models on those tail predicates are remarkably lower than head predicates, as demonstrated in Figure~\ref{sgg-example}~(e).

Prior works have been proposed in recent years to alleviate the bias caused by the long-tail distribution based on causal rules~\cite{tang2020unbiased, li2022devil}, reweighting~\cite{tang2019learning, wang2021seesaw,yu2020cogtree} and resampling strategy~\cite{burnaev2015influence, yan2020pcpl,li2021bipartite} gradually. Nevertheless, these methods still require careful tuning of additional hyper-parameters, such as sampling frequency and category weight. They are sensitive to different architectures and data distributions, which are not flexible for real-world situations. Another alternative way is to increase the number of tail predicates in training. IETrans~\cite{zhang2022fine} uses internal relation correlation to enhance the existing dataset. However, these methods rely on the prior distribution of source data and only work in specific pre-defined conditions. Such a manner based on hand-designed rules covers only limited categories, which is time-consuming and unscalable.

In this paper, we propose a \textbf{C}ross-mod\textbf{a}l predi\textbf{Ca}te b\textbf{o}osting~(\textbf{CaCao}) framework, which leverages the extensive knowledge from the pre-trained language models to enrich the tail predicates of scene graphs in a low-cost and easily scalable way. Our fundamental intuition is that language models gain extensive knowledge about informative relationships from massive text corpus during general sentence pre-training~(\textit{i.e. Large silver \textcolor{blue}{{airp}lane} \textbf{parked outside} an \textcolor{red}{airport} with a \textcolor{blue}{pilot} \textbf{sitting in} it that has \textbf{come back from} a \textcolor{red}{mission}, while the pilot gets some rest.})~\cite{schick2021s, shin2020autoprompt}. 
While the pre-trained language models contain diverse relational knowledge, it is non-trivial to elicit this knowledge from them to scene graph generation. First, there is a significant modality gap in migrating extensive linguistic knowledge into scene graph predicate prediction since such large-scale language models are `blind’ to visual regions. An alternative way is to use vision-language pre-training~(VLP) models. However, VLP models are mainly trained by image-text contrastive learning, lacking the delicate language ability to generate fine-grained predicate category words. Second, a predicate type might correspond to many different linguistic expressions~(\textit{e.g.}, he ``walks through'' / ``is passing through'' / ``passed by'' a street may correspond to the same predicate). Without considering such semantic co-reference phenomenon, the adapted language model for predicate generation can easily collapse to monotonic predictions.

To address the above challenges, we first introduce a novel cross-modal prompt tuning approach, which enables the language model to subtly capture visual context and predict informative predicates as masked language modeling, called the visually-prompted language model. As for semantic co-reference, we further present an adaptive semantic cluster loss for prompt tuning, which models the semantic structures of diverse predicate expressions and adaptively adjusts the distribution to inhibit excessive enhancement of specific predicates during boosting process, thus rendering a diverse and balanced distribution. Moreover, we introduce a fine-grained predicate-boosting strategy to extend the existing dataset with the informative predicates generated by our visually prompted language model. From the comprehensive view of Figure~\ref{sgg-example}~(e), our CaCao can greatly improve the SOTA models' performance in a plug-and-play way, where \textbf{PredCls} of most predicates are consistently increased by 30\% in the purple bar than the blue.

From a more general perspective, our CaCao can not only effectively alleviate the long-tail distribution problem even in large-scale SGG but also generalize to open-world predicates by leveraging the generalizability of human language. Inspired by the impressive zero-shot performance of vision-language pre-training models~\cite{radford2021learning, li2022fine, kim2021vilt}, which utilize the generalizability of human language for zero-shot transfer, we replace the traditional fixed predicate classification layer with category-name embedding and use the diverse predicates generated by our CaCao to learn general and transferable predicate embeddings. Specifically, we propose a novel \textbf{E}ntangled cross-modal \textbf{p}rompt approach for open-world pred\textbf{i}cate s\textbf{c}ene graph generation~(\textbf{Epic}), where the entangled cross-modal prompt alternately tinkers with the predicate representation, making the scene graph model aware of the abstract interactive semantics.

Surprisingly, without using any ground-truth annotations and only with the informative relations generated by our CaCao framework, our Epic achieves competitive performance on the open-world predicate learning problem.

Our main contributions are summarized as follows: 
 \begin{itemize}
     \item We propose a novel \textbf{C}ross-mod\textbf{a}l predi\textbf{Ca}te b\textbf{o}ost-ing~(\textbf{CaCao}) framework, where a visually-prompted language model is learned to enrich the existing dataset with fine-grained predicates in a low-resource and scalable way.

     \item Our CaCao can be applied to SOTA models in a plug-and-play fashion. Experiments over three datasets show steady improvement in standard SGG tasks, demonstrating a promising direction to automatically boosting data by large-scale pre-trained language models rather than time-consuming manual annotation. 
     \item In addition, we introduce \textbf{E}ntangled cross-modal \textbf{p}rompt approach for open-world pred\textbf{i}cate s\textbf{c}ene graph generation~(\textbf{Epic}) to explore the expansibility of CaCao for unseen predicates, and validate its effectiveness with comprehensive experiments.

\end{itemize}

 \begin{figure*}[ht]
\centering
   \includegraphics[width=1.0\linewidth]{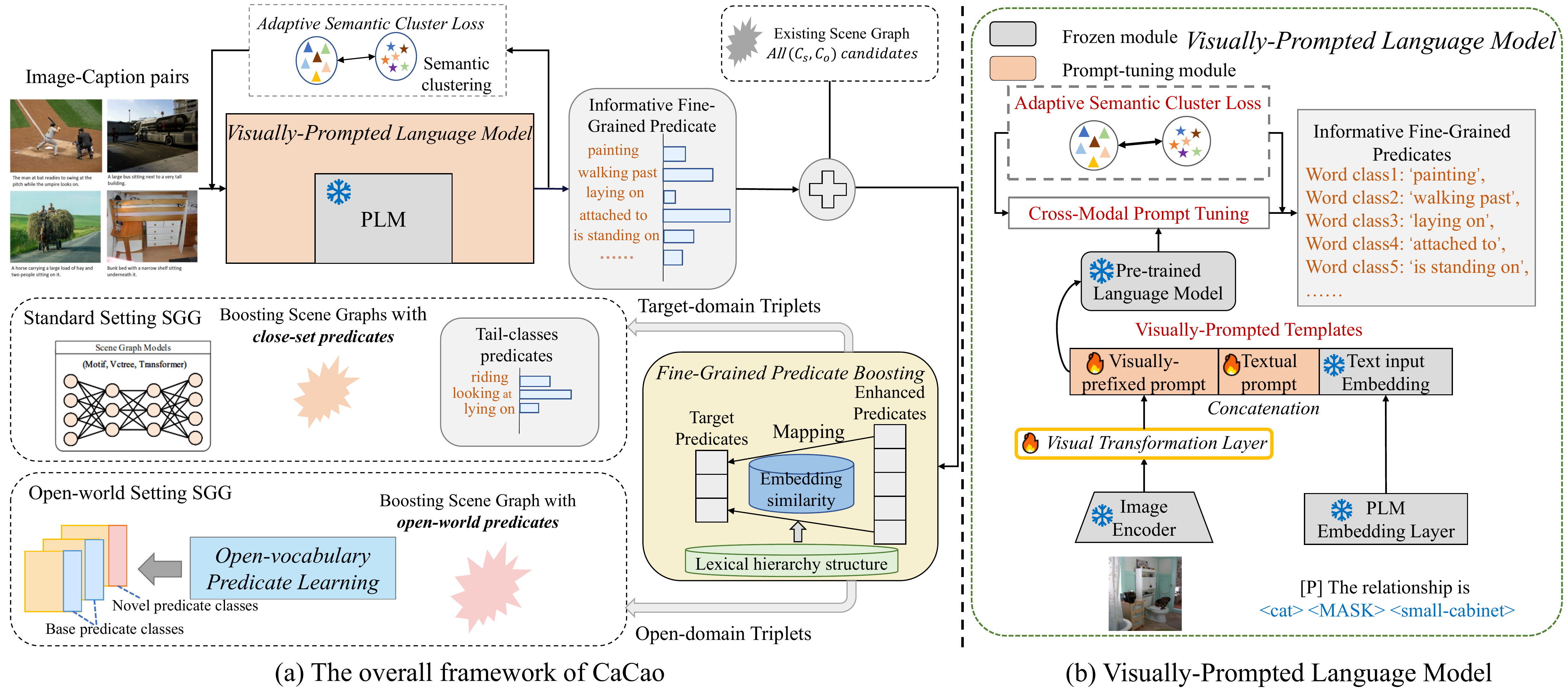}

   \caption{Illustration of our proposed Cross-Modal Predicate Boosting framework. \textbf{Visually-Prompted Language Model} is designed to exploit linguistic knowledge from pre-trained language models and migrate it into scene understanding via visual cues. The right subfigure shows the detail of the visually-prompted language model. \textbf{Fine-Grained Predicate Boosting} uses informative fine-grained predicates to boost the existing scene graph dataset for standard SGG and open-world predicate SGG in a model-agnostic way.}
      \vspace{-0.4cm}
   \label{architecturel}
\end{figure*}

\section{Related Work}
\noindent\textbf{Scene Graph Generation.} Current scene graph generation is still far from practical since it suffers from long-tail distribution of predicates~\cite{zellers2018neural, tang2019learning,chen2019knowledge}. Recently, resampling~\cite{burnaev2015influence,li2021bipartite} and reweighting~\cite{wang2021seesaw,yu2020cogtree} and causal rule-based methods have been proposed to alleviate the biased prediction in the training stage. On the other hand, some approaches aim to balance long-tailed distribution classification following specific class distribution~\cite{zhang2021distribution, desai2021learning}.
Since the predicates in scene graphs are highly relevant to the context, the direct enhancement methods based on class distribution are inapplicable for the balanced scene graph generation.
Hence, \cite{yao2021visual,zhang2022fine} utilize visually relevant relationships from external knowledge bases to address the long-tail predicate problem. However, previous approaches require additional hyper-parameters or hand-designed enhancement rules limited to pre-defined scene graphs. In this work, we propose a predicate-boosting framework that can flexibly enhance SGG datasets with diverse fine-grained predicates.

\noindent\textbf{Language Model Prompting.} Recently, researchers find that large-scale pre-training models contain rich knowledge and exhibit remarkable generalization capabilities for various downstream tasks~\cite{elazar2021measuring, li2023empowering, yu2023interactive, li2022fine}, thereby achieving comparable performance with only little parameter-tuning~\cite{liang2022modular, qin2022deep, li2023gradient, jia2021scaling}. We are also immensely motivated by recent PET work~\cite{schick2021s}, even though it primarily focuses on a semi-supervised situation with many unlabeled instances. FROZEN~\cite{tsimpoukelli2021multimodal} and BLIP-2~\cite{li2023blip} first explore few-shot learning in the multi-modal setting with frozen language models since vision and language can be attended by a unified attention map~\cite{kim2021vilt}. However, these naive prompting methods fail to align complex predicate semantics~(\textsl{i.e.}, ambiguity and co-reference issues) due to their coarse-grained training paradigm. 
We differ from prior works by introducing the first LM with adaptive semantic cluster loss that can distinguish complex predicate semantics from a linguistic perspective, thus efficiently aligning fine-grained visual cues in scene graphs.

\noindent\textbf{Zero-shot Scene Graph Generation.} 
Current zero-shot SGG methods mainly focus on the generalization of relation combination~\cite{lu2016visual,tang2020unbiased,kan2021zero,goel2022not} or only roughly generalize to new categories based on category name similarity~\cite{gkanatsios2020saturation}. However, they fail to effectively handle the intricate and unseen predicates encountered in real-world scenarios.
He~\textit{et al.}~\cite{he2022towards} first introduce open-vocabulary scene graph generation and attempt to predict unseen objects through representation-encoding. But it still cannot transfer to other SGG tasks well because of the enormous cost of dense-caption pre-training. Here we introduce a novel entangled cross-modal prompt to explore the extensibility of CaCao in open-world predicate scene graph generation without costly pre-training.
\section{Cross-Modal Predicate Boosting}
As illustrated in Figure~\ref{architecturel}, our \textbf{C}ross-mod\textbf{a}l predi\textbf{Ca}te b\textbf{o}osting~(\textbf{CaCao}) framework mainly consists of three components: 1) First, the \textit{visually-prompted language model} thoroughly exploits linguistic knowledge from pre-trained language models and migrates it into fine-grained predicates generation. 2) Then, \textit{adaptive semantic cluster loss} is proposed to address the semantic co-reference problem in the visually-prompted language model by diverse predicate expression modeling and adaptive adjustment for predicate enhancement. 3) Finally, \textit{fine-grained predicate boosting} uses these enhanced predicates to alleviate the long-tailed problem of SGG in a model-agnostic way. Furthermore, CaCao can provide various predicates for Epic to achieve open-world SGG. We will elaborate on Epic in Section~\ref{Epic}.
\subsection{Preliminaries}
\noindent\textbf{Scene Graph Generation.} In SGG, we try to locate all objects in the image and predict predicates between them to construct scene graphs. Concretely, given an image $I$, a scene graph $\mathcal{G=(O, R)}$ corresponding to $I$ has a set of objects $\mathcal{O}={(o_i)}_{i=1}^{N_o}$, bounding boxes $\mathcal{B}=(b_i\in{\mathbb{R}^4})$and a set of relationship relationships $\mathcal{R}=(s_i,p_i,o_i)_{i=1}^{N_r}, s_i,o_i\in{\mathcal{O}}$ with different predicate labels $p_i\in\mathcal{P}$, where $N_o$ and $N_r$ are the number of all objects and relationships, respectively.
\subsection{Visually-Prompted Language Model}
\label{VLP-tuning}
 Although several weakly-supervised approaches improve visual relation modeling via specific knowledge bases~\cite{zareian2020weakly, shi2021simple, yao2021visual,zhang2022fine}, they require hand-designed rules and have limited generalization ability. As a result, these methods can only enhance specific predicates and cannot flexibly improve tail predicate prediction in various setups. 
 Thus, we attempt to utilize the linguistic knowledge of pre-trained language models to boost fine-grained predicates in a low-resource way and make language models aware of scenes through visual prompts, as shown in the \textit{visually-prompted language model} module of CaCao in Figure~\ref{architecturel}~(a).

\noindent\textbf{Visually-Prompted Templates.} Due to the modality gap between linguistic knowledge and visual content, language models cannot directly perceive the visual relationships in the scene graph. To better utilize visual semantics, we propose the visually-prompted template containing both visual and textual information, which is designed as \textbf{X}=``[\textit{visually-prefixed prompts}] [P] [SUB][MASK][OBJ]", where [\textit{visually-prefixed prompts}] is an image-conditioned token generated by a transformation layer $h_\theta$ from specific visual features and [P] indicates learnable textual prompt for efficient text prompt engineering.
During training, we feed our visually-prompted templates into frozen language models to predict correct predicates at the masked position and only update the textual prompt [P] together with the parameters $\theta$ in the visual projection layer $h_\theta$.

\noindent\textbf{Cross-Modal Prompt Tuning} aims to predict correct fine-grained predicates at the masked position based on cross-modal contexts from \textbf{X} by optimizing visually-prompted templates. We randomly collect 80k image-caption pairs from the web (\textit{i.e.}, CC3M, COCO caption), which contain nearly 2k categories of predicates but with much noise of simple predicates. We further design heuristic rules (\textit{e.g.}, corpus co-occurrence frequency) to filter out uninformative~(\textit{on, near}) and infrequent~(\textit{kneeling by}) predicates \textbf{automatically} instead of handling them \textbf{manually}. We finally obtain 585 categories of predicates, nearly covering most of the common situations in the real world.
During training, we use a softmax classifier to predict the predicate tokens. Formally, we define $\phi(y_i)$ as a $K$-dimension one-hot label to represent each predicate category $Y_i$~(suppose there are $K$ predicate categories in total).
Given the probability distribution $\psi(y_i|X_i)$ at the masked position for each input $X_i$ and the corresponding predicate label $\phi(y_i)$, we can optimize visually-prompted templates as well as the predicate classifier by the Cross-Entropy Loss as follow:
\vspace*{-0.5\baselineskip} 
\begin{equation}
\mathcal{L} = -\sum_{i=1}^{N_p} {\phi(y_i)log(\psi(y_i|X_i))}~,
\label{virtual}
\vspace*{-0.5\baselineskip} 
\end{equation}
where $N_p$ represents the number of predicates for prompt tuning. Note that we only update the parameters of the visual-linguistic projection layer, as shown in Figure~\ref{architecturel}~(b). 

\subsection{Adaptive Semantic Cluster Loss} 
Although visually-prompted templates partially alleviate semantic ambiguity through instance-conditioned hints, it still suffers from semantic co-reference among predicates, where the same predicate semantic might have multiple linguistic expressions shown in Appendix C. 
Thus, we further design an adaptive semantic clustering loss~(ASCL) to refine diverse predicate semantic expressions through synonym clustering structures and context-aware labels. Additionally, it adaptively suppresses excessively boosted categories based on the distribution of predicates, thus facilitating more various predicate distributions in CaCao.

Specifically, we first represent predicates as the average of the BERT~\cite{devlin2018bert} embedding vectors of its associated triples due to the strong dependency between triplets in complex scenes. We then cluster these predicates using K-means and initialize the number of centroids based on the similarity threshold between each predicate. 
During training, we employ semantic-synonym labels to reduce the penalty for predicates in the same cluster to prevent highly correlated predicates from over-suppressing. The objective is then adjusted by context-aware label and semantic-synonym label as follows:
\begin{equation}
\label{e1}
\begin{split}
&-\min\sum_{i=1}^{N_p}\mathbb{E}_{\boldsymbol{\epsilon}}\big[\underbrace{\phi(y_i)}_{\text {context-aware label }}+ \\
&\underbrace{\sum_{j\in{C_i}}\frac{\epsilon_{i,j}}{|C_i|}\phi(y_j)}_{\text {semantic-synonym label}} \big]log(\psi(y_i|X_i))~,
\end{split}
\end{equation}
where $\epsilon_{i, j}$ is the correlation coefficient between the predicate $y_i$ and other related predicates $y_j$ in its same cluster $C_i$. $|C_i|$ represents the number of predicate categories in it.

Furthermore, we observe that assigned predicate augmentation fails to adequately accommodate the dynamic distribution of predicates, leading to the excessive boosting of some specific predicates that destroys diversity. To address this issue, we set the adaptively re-weighting factor to dynamically adjust the boosting ratio of each predicate based on its proportion during training. We then adjust weights for each category in ASCL as follows:


\begin{align}
\label{e2}
\psi(y_i|X_i) = \frac{e^{z_i}}{\sum_{j=1}^{K} \omega_{ij}e^{z_j}},~\omega_{ij} = \delta\frac{z_j}{z_i}\cdot\frac{n_j}{n_i}~,
\end{align}
where $\{z_i\}_{i=1}^K$ and $\{n_i\}_{i=1}^K$ represent the predicted logit and the initial number of each predicate category $Y_i$, respectively. 
$\omega_{ij}$ denotes the adaptively re-weighting factor concerning dynamic distribution between the target boosted predicate of index $i$ and other predicates of index $j$.
$\delta$ is a hyper-parameter representing prediction margins. When boosting one predicate enough, we will restrain its enhancement by reducing $\omega_{ij}$, guaranteeing the distribution of generated predicates to be balanced and diverse.
\begin{eqnarray}
\label{gar}
    \frac{\partial{\mathcal{L}_i}}{\partial{z_j}}=\frac{(z_j+1)\cdot\frac{\delta n_j}{z_i n_i}e^{z_j}}{\sum_{k=1}^{K} \omega_{ik}e^{z_k}} + \underbrace{\left[\frac{\epsilon_{i,j}e^{z_j}}{\sum_{k=1}^K \omega_{jk}e^{z_k}}-\epsilon_{i,j}\right]}_{\leq 0, j\in{C_i}}~,
\end{eqnarray}
Eq.~\ref{gar} shows the negative gradient for predicate category $Y_j$. The penalty imposed on negative category $Y_j$ is dynamically adjusted as $z_j$ changes. Furthermore, if the negative category $Y_j$ is related to the positive predicate $Y_i$, \textsl{i.e.}, $j\in{C_i}$, we will reduce its punishment to encourage the diversity of CaCao. Finally, it results in an adaptively boosting process to promote predicate diversity in CaCao.

\subsection{Fine-Grained Predicate Boosting}
\label{Knowledge relabeling}
Although we obtain abundant fine-grained predicates from CaCao, it is not straightforward to directly boost them into the target scene graph due to category inconsistencies with the predicates in the target scene graph. To  address this limitation, we propose a fine-grained predicate boosting stage to effectively map open-world predicates to target categories, guaranteeing the smooth alignment of the enhanced predicates with the target scene graph.

Specifically, we establish a simple hierarchy structure of target predicate categories based on lexical analysis and map fine-grained predicates to the target category at each level by cosine similarity of triplet-level embedding. We then select the least frequent category from the mapped candidate target predicates as the final predicate. Note that we only boost unlabeled object pairs that overlap in the scene graph to preserve the original semantics. We will explore more complex structures in the future.

Given the existing scene graph dataset with $\left| \mathcal{N} \right|$ labeled samples, our CaCao can generate extra training data $\mathcal{D}$ automatically in a low-resource way and flexibly extend the current dataset  by fine-grained predicate boosting. Finally, we retrain the refined SGG models with enhanced data $\mathcal{\hat{N}}=(\mathcal{N},\mathcal{D})$ for a more balanced prediction. We then formulate the learning problem as follows:
 \vspace*{-0.4\baselineskip}
\begin{equation}
    \min\limits_{\theta}{\frac{1}{\left| \mathcal{\hat{N}} \right|}\sum_{i=1}^{\left| \mathcal{\hat{N}} \right|}{L(N_i;\theta)}}~,
\vspace*{-0.4\baselineskip}
\end{equation}
where $L(N_i;\theta)$ denotes the loss function of the learning procedure during the standard scene graph generation. 

\begin{figure}[!t]
  \centering
  
   \includegraphics[width=1.0\linewidth]{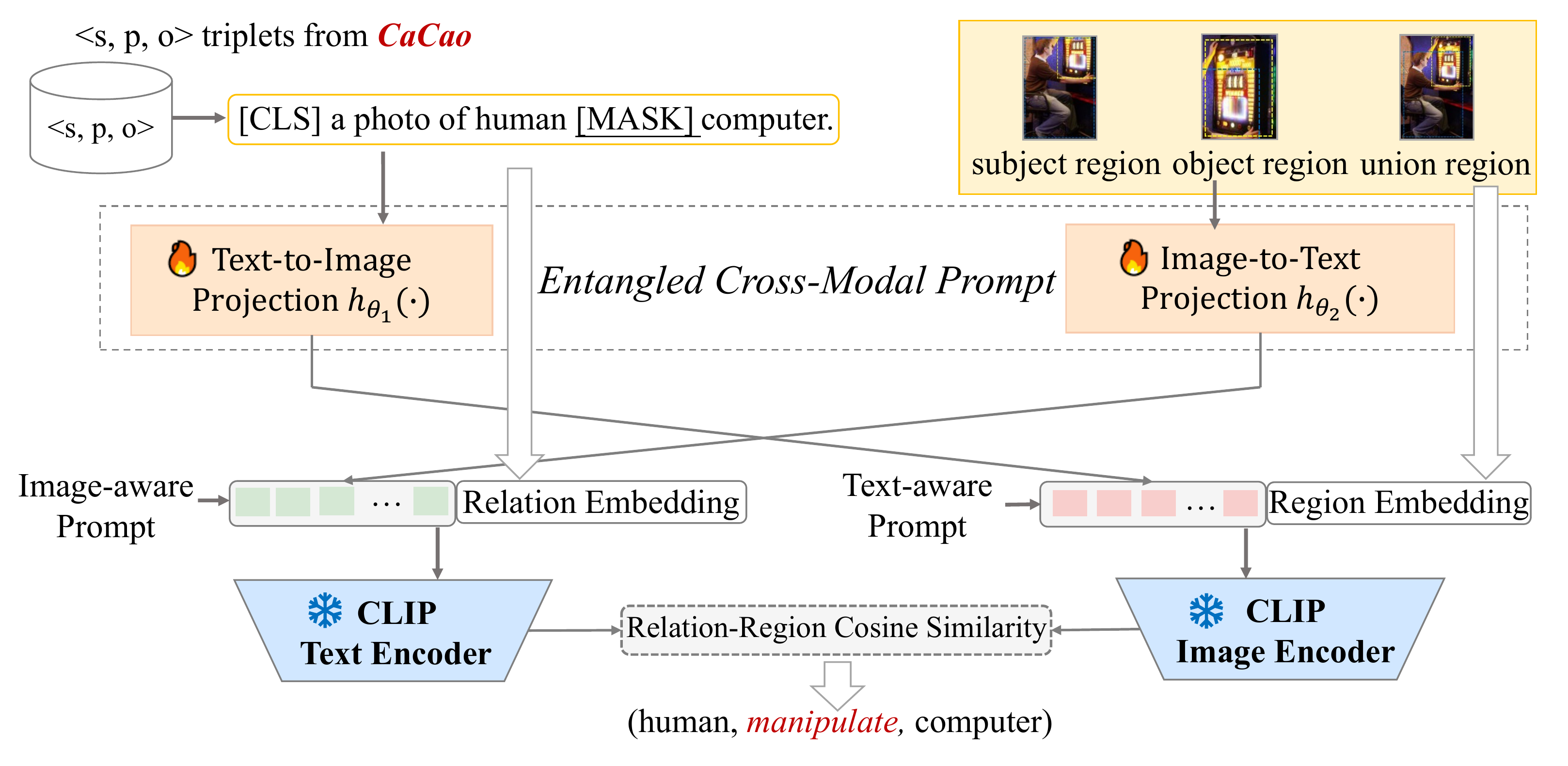}
   \caption{Illustration of our proposed \textbf{E}ntangled cross-modal \textbf{p}rompt approach for open-world pred\textbf{i}cate s\textbf{c}ene graph generation~(\textbf{Epic}). It guides the model to learn the unified embedding of predicates by two complementary prompts in an associative way.}
    \vspace{-0.3cm}
   \label{entangledprompt}
\end{figure}

\section{Open-World Predicate SGG}
\label{Epic}
Since CaCao can generate fine-grained predicates, it can provide extra unseen data for open-world generalization.
However, open-world predicate SGG has two extra challenges: (a) understanding multi-level semantics of images and triplets; (b) aligning novel predicate semantics into visual and textual contexts.
To this end, we propose a novel \textbf{E}ntangled cross-modal \textbf{p}rompt approach for open-world pred\textbf{i}cate s\textbf{c}ene graph generation~(\textbf{Epic}). With the help of Epic, we can fully exploit the potential of CaCao and extend it into open-world predicate SGG, as shown in Figure~\ref{entangledprompt}.

\noindent\textbf{Open-World Predicate SGG Backbone.} A straightforward way to predict unseen classes in an open world is replacing the fixed classifier with unified embeddings~\cite{radford2021learning}. Let $x$ be the region embedding generated by the visual encoder and $\{p_i\}_{i=1}^{K}$ be a set of relation embeddings produced by the text encoder, $p_*$ is the embedding of the correct predicate. The loss for open-world predicate SGG is then,
\begin{equation}
\label{cl_loss}
    \mathcal{L}_{open} = -{\rm log}\frac{{exp}(sim({x}, {p_*})/\tau)}{\sum_{i=1}^{K}{{\rm exp}(sim({x},{p_i})/\tau)}}~,
\end{equation}
where $p_*$ is the matched relation embedding and $sim(\cdot,\cdot)$ denotes the cosine similarity. $\tau$ is a temperature parameter. 

\noindent\textbf{Entangled Cross-Modal Prompt.} 
Moreover, we notice that predicate semantics and image regions are closely related to visual and textual contexts. For example, ``man on chair" and ``shirt on chair" represent different semantics even though they correspond to the same image region in Figure~\ref{sgg-example}~(a); ``man" and ``horse"  may correspond to different predicates ``riding" or ``holding" in different visual contexts. Inspired by the remarkable performance of prompts~\cite{liu2022p,li2021prefix,zhou2022learning}, we introduce entangled cross-modal prompts for text encoder and image encoder to alleviate the above problems. Let $h_{\theta_1}(\cdot)$, $h_{\theta_2}(\cdot)$ represent the the text-to-image and image-to-text projection, respectively in Figure~\ref{entangledprompt}. The predicate probability is then computed as:
\begin{equation}
\label{cl_loss_ours}
    P~(p^*|{x}) = \frac{{\rm exp}~(sim~(f_x~({p_*}), {t}_{p_*}~({x}))/\tau)}{\sum_{i=1}^{K}{{\rm exp}~(sim~({f}_x~({p_i}), {t}_{p_i}~({x}))/\tau)}}~,
\end{equation}
where $f_x~(p_i)$ is conditional region embedding based on text-aware prompt $h_{\theta_1}(p_i)$ and $t_{p_i}~(x)$ is conditional relation embedding based on image-aware prompts $h_{\theta_2}(x)$. During training, we only update the projection parameters $~(\theta_1;\theta_2)$ to preserve the pre-trained language-vision model’s capability for open-world predicate generalization.

\begin{table*}[]
\resizebox{\textwidth}{!}{
\begin{tabular}{ccp{2cm}cccccc}
\hline
\toprule
&\multicolumn{1}{c}{\multirow{2}{*}{Model Type}} & \multicolumn{1}{c}{\multirow{2}{*}{Methods}} & 
\multicolumn{2}{c}{Predicate Classification} & \multicolumn{2}{c}{Scene Graph Classification} & \multicolumn{2}{c}{Scene Graph Detection} \\ 
 & & & Tail-R@20/50/100 $\uparrow$ & mR@20/50/100 $\uparrow$&Tail-R@20/50/100 $\uparrow$ & mR@20/50/100 $\uparrow$ &Tail-R@20/50/100 $\uparrow$ & mR@20/50/100 $\uparrow$
 \\ \hline
&\multicolumn{1}{c}{\multirow{4}{*}{Specific}} & \multicolumn{1}{l}{BGNN~\cite{li2021bipartite}} & -/- & - / 30.4 / 32.9 & -/- & - / 14.3 / 16.5 &  -/- & - / 10.7 / 12.6  \\
&& \multicolumn{1}{l}{PCPL~\cite{yan2020pcpl}} & -/- & - / 35.2 / 37.8 & -/- & - / 18.6 / 19.6 & -/- & - / 9.5 / 11.7  \\ 
&& \multicolumn{1}{l}{SVRP~\cite{he2022towards}} & -/- &  - / 24.3 / 25.3 & -/- & - / 12.5 / 15.3 & -/- &  - / 10.5 / 12.8  \\
&& \multicolumn{1}{l}{DT2-ACBS~\cite{desai2021learning}} & -/- & 27.4 / 35.9 / 39.7 & -/- & 18.7 / 24.8 / 27.5 &  -/- & 16.7 / 22.0 / 24.4 \\
\hline\hline

&\multicolumn{1}{c}{\multirow{2}{*}{One-stage}} & \multicolumn{1}{l}{SSRCNN~\cite{teng2022structured}} & -/- & -/- & -/- & -/- & 10.4 / 16.3 / 19.1 & 13.7 / 18.6 / 22.5  \\
&&\multicolumn{1}{l}{\quad\textbf{+CaCao~(ours)}} & -/- & -/-  & -/- & -/- &  \textbf{13.6 / 18.0 / 21.2} & \textbf{14.1 / 18.7 / 23.1} \\
\hline\hline

\multirow{21}{*}{\rotatebox[]{90}{Model-Agnostic strategy}}& & \multicolumn{1}{l}{Motif~\cite{zellers2018neural}} & 10.2 / 13.3 / 14.4 &  12.1 / 15.2 / 16.2 &5.8 / 6.8 / 7.3  & 7.2 / 8.7 / 9.3 & 4.8 / 6.0 / 7.3  & 5.1 / 6.5 / 7.8   \\
&\multicolumn{1}{c}{Resample} & \multicolumn{1}{l}{\quad+Resample~\cite{burnaev2015influence}} & -/- &  14.7 / 18.5 / 20.0 & -/- & 9.1 / 11.0 / 11.8 & -/- &  5.9 / 8.2 / 9.7 \\ 
&\multicolumn{1}{c}{\multirow{2}{*}{Reweight}}& \multicolumn{1}{l}{\quad+Reweight~\cite{wang2021seesaw}} & 16.7 / 26.3 / 31.0 & 18.8 / 28.1 / 33.7 & 8.9 / 11.8 / 14.1 &  10.7 / 15.6 / 18.3 & 8.6 / 12.1 / 14.6 & 7.2 / 10.5 / 13.2 \\
&&\multicolumn{1}{l}{\quad+FGPL~\cite{lyu2022fine}} & 26.7 / 33.3 / 35.7  & 24.3 / 33.0 / 37.5 &  16.8 / 19.1 / 19.9 & 17.1 / 21.3 / 22.5 & 12.4 / 16.5 / 19.3 & 11.1 / 15.4 / 18.2\\
&\multicolumn{1}{c}{Causal Rule}& \multicolumn{1}{l}{\quad+TDE~\cite{tang2020unbiased}} & -/- & 18.5 / 25.5 / 29.1 & -/-
& 9.8 / 13.1 / 14.9 & -/- & 5.8 / 8.2 / 9.8 \\ \cline{2-9}
&\multicolumn{1}{c}{\multirow{5}{*}{Data Enhancement}}& \multicolumn{1}{l}{\quad+Only Caption Relations} & 16.7 / 20.5 / 21.8  &  15.2 / 19.8 / 21.2 & 8.1 / 9.6 / 10.1 & 8.0 / 9.8 / 10.5  & 5.3 / 7.7 / 9.4 & 6.0 / 8.2 / 10.0  \\
&&\multicolumn{1}{l}{\quad+VisualDS~\cite{yao2021visual}} & 11.3 / 14.5 / 16.3  & 13.1 / 16.1 / 17.5 &  5.9 / 7.0 / 8.3  &  7.6 / 9.3 / 9.9  & 5.1 / 6.8 / 7.8 & 5.4 / 7.0 / 8.3 \\
&&\multicolumn{1}{l}{\quad+DLFE~\cite{chiou2021recovering}} & -/- & 22.1 / 26.9 / 28.8  &  -/- & 12.8 / 15.2 / 15.9  & -/- & 8.6 / 11.7 / 13.8 \\ 
&&\multicolumn{1}{l}{\quad+IETrans~\cite{zhang2022fine}} & 27.3 / 31.3 / 33.2 & 30.2 / 35.8 / 39.1 &  13.5 / 15.5 / 16.1 & 18.2 / 21.5 / 22.8 & 9.2 / 12.3 / 14.3 & 12.0 / 15.5 / 18.0 \\
&&\multicolumn{1}{l}{\quad\textbf{+CaCao~(ours)}} & \textbf{31.4 / 36.1 / 37.6} & \textbf{30.9 / 37.1 / 38.9}  & \textbf{17.3 / 19.7 / 20.5} & \textbf{20.4 / 23.3 / 24.4} &  \textbf{13.9 / 18.4 / 21.6} & \textbf{12.6 / 17.1 / 20.0}\\ 
\cline{2-9}\specialrule{0em}{1.5pt}{1.5pt}\cline{2-9}

&&\multicolumn{1}{l}{VCTree~\cite{tang2019learning}} & 9.9 / 13.0 / 14.0 & 11.7 / 14.9 / 16.1 & 6.2 / 7.4 / 7.9& 9.1 / 11.3 / 12.0 & 4.3 / 6.1 / 7.2 & 5.2 / 7.1 / 8.3\\
&\multicolumn{1}{c}{\multirow{2}{*}{Reweight}}&\multicolumn{1}{l}{\quad+Reweight~\cite{wang2021seesaw}} & 23.9 / 30.7 / 33.7 & 19.4 / 29.6 / 35.3 & 12.2 / 14.9 / 16.1 & 13.7 / 19.9 / 23.5 & 8.4 / 12.2 / 14.7 &  7.0 / 10.5 / 13.1 \\
&&\multicolumn{1}{l}{\quad+FGPL~\cite{lyu2022fine}} & 32.2 / 36.8 / 38.2 & 30.8 / 37.5 / 40.2 &  23.5 / 26.5 / 27.5 &  21.9 / 26.2 / 27.6 & 13.5 / 17.4 / 20.4 &  \textbf{11.9} / 16.2 / 19.1\\
&\multicolumn{1}{c}{Causal Rule}&\multicolumn{1}{l}{\quad+TDE~\cite{tang2020unbiased}} & -/- &  18.4 / 25.4 / 28.7 & -/-
&  8.9 / 12.2 / 14.0 & -/- &  6.9 / 9.3 / 11.1 \\ \cline{2-9}
&\multicolumn{1}{c}{\multirow{4}{*}{Data Enhancement}}&\multicolumn{1}{l}{\quad+Only Caption Relations} & 16.2 / 20.3 / 21.7 & 14.7 / 19.3 / 20.9 & 8.0 / 9.8 / 10.4 & 8.2 / 10.1 / 10.8 & 6.0 / 8.0 / 9.7 & 5.5 / 7.8 / 9.5 \\
&&\multicolumn{1}{l}{\quad+DLFE~\cite{chiou2021recovering}} & -/- & 20.8 / 25.3 / 27.1 &  -/- & 15.8 / 18.9 / 20.0  & -/- & 8.6 / 11.7 / 13.8 \\ 
&&\multicolumn{1}{l}{\quad+IETrans~\cite{zhang2022fine}} & 27.3 / 31.6 / 33.0 & 31.7 / 37.0 / 39.7 & 11.6 / 13.6 / 14.3 & 18.2 / 19.9 / 21.8  & 9.0 / 11.8 / 13.7 & 9.8 / 12.0 / 14.9 \\
&&\multicolumn{1}{l}{\quad\textbf{+CaCao~(ours)}} & \textbf{33.1 / 37.5 / 38.9} & \textbf{33.8 / 39.0 / 40.8}  & \textbf{23.8 / 27.2 / 28.2} & \textbf{23.8 / 27.5 / 28.7} &  \textbf{14.6 / 19.4 / 22.6} & 11.8 \textbf{/ 16.4 / 19.1}  \\ 
\cline{2-9}\specialrule{0em}{1.5pt}{1.5pt}\cline{2-9}

&&\multicolumn{1}{l}{Transformer~\cite{tang2020unbiased}} & 10.8 / 13.5 / 14.6 & 12.4 / 16.3 / 17.6 & 8.8 / 10.3 / 11.8 & 8.7 / 10.1 / 10.7 & 5.3 / 7.3 / 8.8 & 5.8 / 8.1 / 9.6  \\ 
&\multicolumn{1}{c}{\multirow{2}{*}{Reweight}}&\multicolumn{1}{l}{\quad+Reweight~\cite{wang2021seesaw}} & 19.9 / 26.0 / 28.4 &  19.5 / 28.6 / 34.4 & 9.5 / 12.6 / 13.4 & 11.9 / 17.2 / 20.7 & 7.0 / 10.3 / 12.4 &  8.1 / 11.5 / 14.9 \\ 
&&\multicolumn{1}{l}{\quad+FGPL~\cite{lyu2022fine}} & 26.6 / 33.6 / 36.0 & 27.5 / 36.4 / 40.3 &  17.0 / 19.9 / 20.1 &  19.2 / 22.6 / 24.0 & 13.1 / 17.0 / 19.8 &  13.2 / 17.4 / 20.3\\
\cline{2-9}
&\multicolumn{1}{c}{\multirow{3}{*}{Data Enhancement}}&\multicolumn{1}{l}{\quad+Only Caption Relations} & 16.1 / 19.4 / 20.8 & 15.0 / 19.3 / 20.9 & 8.3 / 9.9 / 10.5 & 8.6 / 10.6 / 11.2 & 6.4 / 8.9 / 10.6 & 6.0 / 8.4 / 10.4 \\
&&\multicolumn{1}{l}{\quad+IETrans~\cite{zhang2022fine}} & 27.5 / 32.0 / 33.7& 29.1 / 35.0 / 38.0 & 14.1 / 16.2 / 16.7  & 17.9 / 20.8 / 22.3  & 11.6 / 14.9 / 17.6 & 11.7 / 15.0 / 18.1 \\
&&\multicolumn{1}{l}{\quad\textbf{+CaCao~(ours)}} & \textbf{31.7 / 35.7 / 37.0} & \textbf{36.2 / 41.7 / 43.7}  & \textbf{19.0 / 22.2 / 23.3} & \textbf{21.1 / 24.0 / 25.0} &  \textbf{14.1 / 18.7 / 21.9} & \textbf{13.5 / 18.3 / 22.1} \\
\bottomrule
\end{tabular}
}
\vspace{0.5mm}
\caption{Performance~(\%) of our method \textbf{CaCao} and other baselines with different model types on the VG-50 dataset. }
\label{base-result}
\vspace{-3mm}
\end{table*}
\begin{table}[ht]
\setlength\tabcolsep{1pt}
\resizebox{0.475\textwidth}{!}{
\begin{tabular}{ccccc}
\hline
\toprule
&Model &  PredCls mR@50/100& SGCls mR@50/100& SGDet mR@50/100\\ \hline
\multirow{6}{*}{\rotatebox[]{90}{GQA-200}}&Motif~\cite{zellers2018neural} & 16.4 / 17.1 & 8.2 / 8.6 & 6.4 / 7.7\\
&Motif + GCL\cite{dong2022stacked} & 36.7 / 38.1 & 17.3 / 18.1 & 16.8 / 18.8\\
&\textbf{Motif + CaCao~(ours)} & \textbf{37.5 / 40.5} & \textbf{19.6 / 21.9} & \textbf{17.8 / 19.6}\\
&Transformer\cite{tang2020unbiased} & 17.5 / 18.7 & 8.5 / 9.0 & 6.6 / 7.8 \\
&Transformer + GCL\cite{dong2022stacked} & \textbf{35.6} / 36.7 & 17.8 / 18.3 & 16.6 / 18.1 \\
&\textbf{Transformer + CaCao~(ours)} & 34.8 / \textbf{36.9} & \textbf{19.3 / 20.1} & \textbf{18.8 / 19.1}\\ \hline
\multirow{4}{*}{\rotatebox[]{90}{VG-1800}}&BGNN~\cite{li2021bipartite} & 1.3 / 2.4 & 0.8 / 1.4 & 0.5 / 0.9 \\ 
&Motif~\cite{zellers2018neural} & 1.7 / 2.6  &  0.9 / 1.9 & 0.6 / 1.1\\
&Motif + IETrans~\cite{zhang2022fine} & 5.1 / 8.4 & 3.6 / 5.2 & 3.1 / 4.3\\
&\textbf{Motif + CaCao~(ours)} & \textbf{10.0 / 10.8} & \textbf{4.6 / 6.3} & \textbf{4.1 / 6.2}\\ 
\bottomrule
\end{tabular}
}
\vspace{0.3mm}
\caption{Comparisons with our CaCao and other baseline methods on large-scale SGG datasets.}
\vspace{-1mm}
\label{largescale-result}

\end{table} 

\begin{table}[ht]
\setlength\tabcolsep{1pt}
\resizebox{0.475\textwidth}{!}{
\begin{tabular}{llcccc}
\hline
\toprule
 &\multicolumn{1}{c}{\multirow{2}{*}{Methods}} &\multicolumn{1}{c}{\multirow{2}{*}{Datasets}}&\multicolumn{3}{c}{Predicate Classification}  \\
 && &base R@50/100 & novel R@50/100 & total mR@50/100
 \\ \hline\hline
& \multicolumn{1}{c}{\multirow{3}{*}{Backbone w/o Epic~\cite{radford2021learning}}} & VG & 17.6 / 21.1  & 6.4 / 8.7 & 8.5 / 9.7\\
 & & CaCao& 17.4 / 20.4 & 7.2 / 9.2 & 8.1 / 10.4 \\
 && VG+CaCao& 17.5 / 20.9 & 11.2 / 15.8 & 13.6 / 17.7  \\
\hline
 \multirow{5}{*}{\rotatebox[]{90}{Ablations}} & \multicolumn{1}{c}{\multirow{3}{*}{\textbf{Epic}}} & VG& 22.6 / 27.2 & 7.4 / 9.7 & 10.3 / 12.6\\
 && CaCao&23.1 / 30.8 & 9.7 / 12.1 & 14.2 / 18.2   \\
 && VG+CaCao & \textbf{28.3 / 31.1} & \textbf{13.9 / 18.3}  & \textbf{16.5 / 21.8}  \\
\cline{2-6}
 &\quad w/o text-aware prompt & VG+CaCao& 16.8 / 23.1 & 12.5 / 13.9 & 13.1 / 15.4\\
 &\quad w/o vision-aware prompt & VG+CaCao&18.5 / 24.9 & 10.1 / 12.7 & 11.2 / 14.1  \\
\bottomrule
\end{tabular}
}
\vspace{0.3mm}
\caption{Performance~(\%) of our \textbf{Epic} and the backbone without Epic for open-world settings on different datasets. VG denotes the VG-50 dataset with the open-world split, VG+CaCao represents the enhanced dataset with our CaCao framework and CaCao means only use CaCao's predicates for unsupervised settings.}
\label{zs-result}
\vspace{-5.8mm}
\end{table}

\section{Experiment}

\subsection{Dataset and Evaluation Settings}
\noindent\textbf{Datasets.} We evaluate our proposed method for scene graph generation on the popular VG-50 benchmark similar to previous works~\cite{krishna2017visual, xu2017scene,tang2020unbiased,tang2019learning,zellers2018neural}, which consists of 50 predicate classes and 150 object classes. Furthermore, we explore more challenging datasets~(\textit{i.e.} GQA-200~\cite{hudson2019gqa, knyazev2021generative}, VG-1800~\cite{zhang2022fine}) where predicates are more diverse to
validate CaCao's generalization ability in large-scale scenarios.

\noindent\textbf{Data Split.}
For the standard SGG setting, we adopt a widely used data split following previous works~\cite{tang2020unbiased, zellers2018neural, knyazev2021generative} and expand to large-scale SGG datasets. we divide the dataset into 70\% training set, 30\% testing set, and additional 5k images for parameter tuning.
For the open-world predicate SGG setting, we first establish the related dependencies from Chen \textit{et al}~\cite{chen2019scene}. We then randomly select 70\% classes from each predicate level and assign them into the base set for training and the rest 30\% classes that contain rare predicates~(\textit{e.g.}, painted on, flying in) into the novel set for evaluation similar to other zero-shot tasks~\cite{bansal2018zero, kan2021zero,he2022towards}. To avoid disclosure of the unseen predicates, we remove all relations that contain novel predicates in the training set. 

\noindent\textbf{Evaluation and Metrics.} 
Following recent works~\cite{zhang2022fine, lyu2022fine}, we evaluate our model on three widely used SGG tasks: PredCls, SGCls, and SGDet. Since the Recall@K of all predicates could be easily affected by biased distribution, it cannot precisely evaluate models' performance on long-tail distribution SGG. 
Thus, we use Mean Recall@K~(\textbf{mR@K}) to evaluate the performance of SGG models on the whole category set. 
We further introduce a detailed metric \textbf{Tail-R@K}~(Recall@K among tail 50\% predicates) to better assess those tail predicates, as these predicates typically provide more information for image understanding.
Besides, we use Recall@K of base predicates, novel predicates, and mean Recall@K of total predicates to evaluate the generalization ability of our method on open-world predicate SGG.

\subsection{Implementation Details}\label{5}
\noindent\textbf{Visually-Prompted Language Model.} We use ViT~\cite{dosovitskiy2020image} as the image encoder and set a transformer layer with the 768 embedding size to obtain visually-prefixed prompts. We set the length of visual prompts as 50 and set 10 learnable tokens as textual prompts [\textit{P}] for textual alignment. We use BERT~\cite{devlin2018bert} as the language model to predict target predicates and train the model for 15 epochs with batch size of 32. We use AdamW~\cite{loshchilov2017decoupled} to optimize the model and set the basic learning rate as 3e-4 with a weight delay of 0.0004. Furthermore, the prediction margin $\delta$ is set as 9.0.

\noindent\textbf{Object Detector.} Following previous works\cite{tang2020unbiased, tang2019learning, li2021bipartite}, we use a pre-trained Faster R-CNN~\cite{ren2015faster} with ResNet-101-FPN~\cite{he2016deep} as our backbone and train it on VG-50 dataset with SGD as the optimizer. We then fix the parameters of the object detector during standard SGG training.  

\noindent\textbf{Scene Graph Generation.} We follow almost the framework of the SOTA unbiased SGG method~\cite{lyu2022fine}, the only difference is that we integrate the enhanced triplets derived from CaCao into SGG training, thereby directing more attention towards tail predicates without any extra costs. Following~\cite{tang2020unbiased}, SGG models are trained with Cross-Entropy Loss and SGD optimizer by initial learning rate as 1e-3, and batch size as 16. Besides, we train SGG models with 16000 batch iterations for all sub-tasks. For the GQA-200 and VG-1800 datasets, we adjust the training batch iterations to 80000 for further training in large-scale SGG.

For open-world predicate SGG, we use CLIP~\cite{radford2021learning} as the backbone to obtain region embedding and predicate embedding. The text-to-vision and vision-to-text projections are two-layer structures~(Linear-ReLU-Linear) with the model dimension $d=512$ to get the conditional prompts. We set the length of the vision-aware prompt to 4 and the length of the text-aware prompt to 2. Then we use the InfoNCE~\cite{oord2018representation} loss and set the temperature as 0.9 with the batch size of 4 to learn the representation of predicate categories.
\subsection{Comparison with State of the Arts}
We report the results of our CaCao and other general SGG models for the VG-50 benchmark shown in Table~\ref{base-result}. Based on the observation of experimental results, we have summarized the following conclusions:

\textbf{Our CaCao framework can be flexibly equipped to different baseline models.} We incorporate our CaCao into three backbone models for evaluation, including Motif~\cite{zellers2018neural}, VCTree~\cite{tang2019learning}, and Transformer~\cite{tang2020unbiased}. Despite the model diversity, our CaCao can consistently improve all baseline models’ mR@K performance for all tasks that Motif+CaCao~(38.9\% \textit{v.s.} 16.2\%), VCTree+CaCao~(40.8\% \textit{v.s.} 16.1\%) and Transformer+CaCao~(43.7\% \textit{v.s.} 17.6\%) for PredCls. Also, we obtain similar performance improvements for SGCls and SGDet. Besides, we compare the ablation methods which directly extract raw relation triplets from captions~(\textit{i.e.} only Caption Relations in Tab.~\ref{base-result}). Notably, our method Transformer+CaCao significantly surpasses the ablation method by 23.8\% in mR@20 of PredCls, demonstrating that the gain power of CaCao is mainly derived from linguistic knowledge in PLM instead of extra collected data. Conversely, the triplets directly extracted from image captions are incomplete that only describe general semantics or partial visual relationships.

\textbf{Compared with other model-agnostic methods, our CaCao outperforms all of them in both Tail-R@K and mR@K.} Specifically, CaCao exceeds the SOTA of data enhancement models IETrans~\cite{zhang2022fine} for all three backbones with consistent improvements as 3.9\%, 5.8\%, and 3.6\% on Tail-R@20 for PredCls and 0.7\%, 2.1\%, and 7.1\% on mR@20 for Predcls. It shows that our CaCao can generate high-quality informative predicates to mitigate the long-tail distribution problem, which is conducive to fine-grained scene graph generation. It is worth noting that even when compared to SOTA methods of different model types, such as FGPL~\cite{lyu2022fine}, Motif+CaCao, VCTree+CaCao, and Transformer+CaCao still achieve significant improvements by 6.6\%, 3.0\%, and 8.7\% on mR@20 for PredCls. Besides, our CaCao can also integrate with one-stage methods~(\textit{e.g.}, SSRCNN~\cite{teng2022structured}) and achieve better performance.

\begin{figure}[ht]\vspace{-0.25cm}
    \centering
   \includegraphics[width=1.0\linewidth]{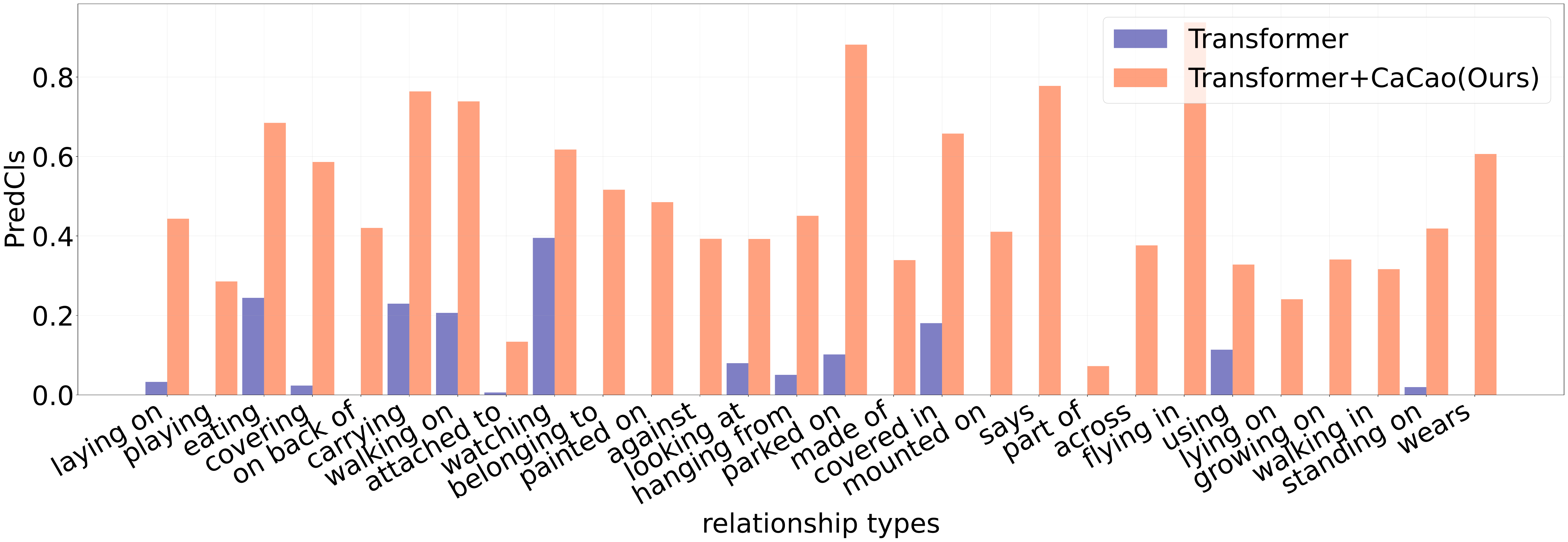}
   \caption{Diverse fine-grained predicates performance comparison between base Transformer~\cite{tang2020unbiased} and our enhanced Transformer+CaCao on the VG-50 dataset.}
   \label{reuslts-compare}
   \vspace{-0.45cm}
\end{figure}
\begin{figure}
\centering
\includegraphics[width=1.\linewidth]{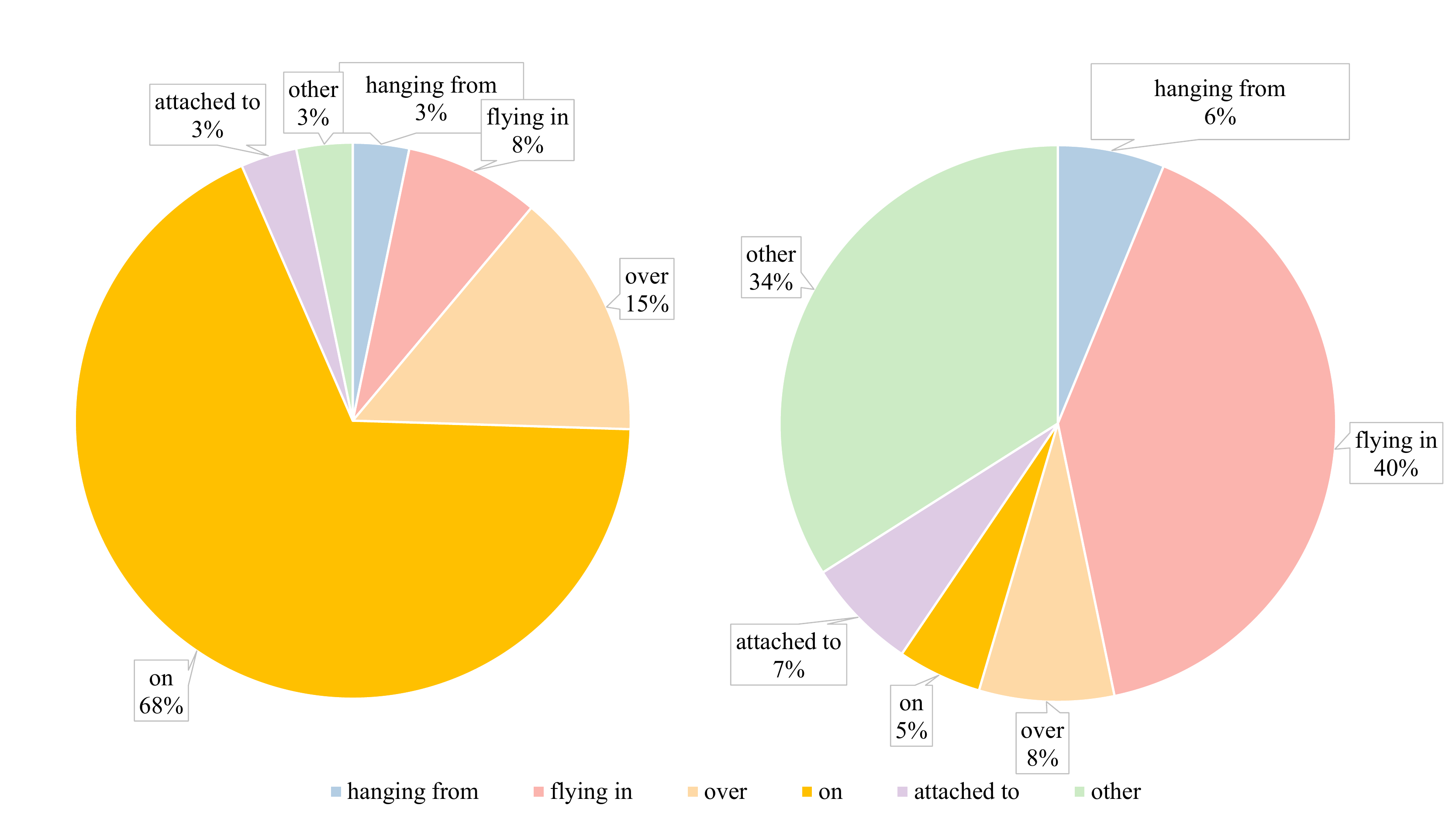}
\caption{The prediction distribution of CaCao on the fine-grained predicate `\textit{flying in}'. The left pie chart shows the distribution by Transformer~\cite{tang2020unbiased} and the right pie chart shows the prediction distribution of various predicates by Transformer+CaCao.}
\label{visual-distribution}
\end{figure}

\textbf{Our method can distinguish fine-grained predicates and achieve a large margin of improvements on these predicate predictions.} 
Notably, our modal-agnostic approach can also achieve competitive performance compared with strong specific baselines~(\textit{e.g.}, 43.7\% \textit{v.s.} 39.7\% on mR@100 for PredCls), demonstrating the superiority of our proposed model. For an intuitive illustration of CaCao's discriminatory power among hard-to-distinguish predicates, we visualize the PredCls results of fine-grained predicates as shown in Figure~\ref{reuslts-compare}. We observe that Transformer+CaCao obtains overall improvement on most predicates. One possible reason is that CaCao has been exposed to various informative predicates, strengthening its discriminatory power against fine-grained predicates. 
Qualitatively, we further visualize the prediction results of our Transformer+CaCao compared with its baseline model Transformer~\cite{tang2020unbiased}, shown in Figure~\ref{visual-distribution}. In the case of Transformer+CaCao, we observe a substantial improvement in the predicted ratio for the correct predicate `\textit{flying in}'~(8\% $\rightarrow$ 40\%). This result demonstrates the capability of Transformer+CaCao to effectively distinguish fine-grained predicates, as opposed to roughly predicting head predicates~(\textsl{i.e.}, on, in).

\begin{figure*}[ht]
    \centering
   \includegraphics[width=0.98\linewidth]{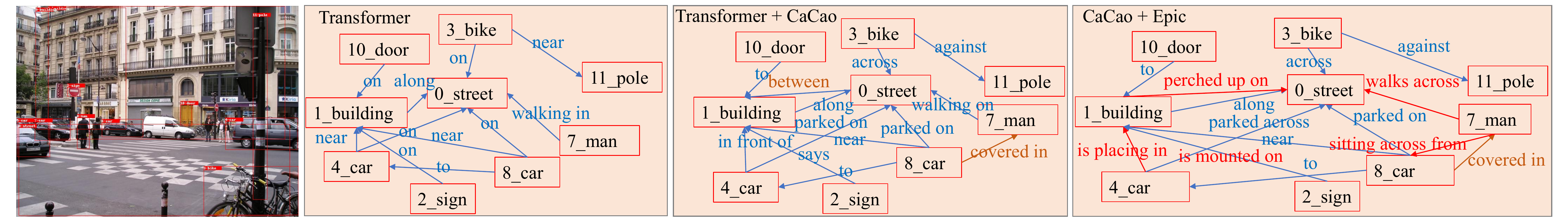}
   
   \caption{Visualization of base Transformer model~\cite{tang2020unbiased}, Transformer equipped with our CaCao framework for predicate enhancement and our Epic equipped with CaCao framework for open-world predicate SGG.}
    \vspace{-0.4cm}
   \label{visualization}
\end{figure*}
\subsection{Generalization to Large-Scale SGG}
Table~\ref{largescale-result} summarizes the results of our CaCao and other baselines on large-scale datasets. Overall, our method can successfully generalize to more challenging datasets. Notably, simply resampling~(\textit{i.e.} BGNN~\cite{li2021bipartite}) can not work well in such exacerbated scenarios, where much more predicates have less than 10 samples. In contrast, our CaCao utilizes abundant corpus knowledge to balance diverse tail predicates and surpasses other baselines for almost large-scale SGG tasks. For quantitative comparison, our CaCao can obtain consistent improvement as 8.2\%, 4.4\%, and 5.1\% on mR@100 for PredCls, SGCls, and SGDet in VG-1800 and largely enhance the unbiased predictions in GQA-200~(\textit{e.g.}, 11.9\% improvement with Motif~\cite{zellers2018neural} and 11.7\% improvement with Transformer~\cite{tang2020unbiased} on SGDet mR@100).

\subsection{Expansibility to Open-World Predicate SGG}
Inspired by the abundant fine-grained predicates produced by CaCao, we also validate our CaCao with Epic on the open-world setting for the base, novel, and total PredCls tasks to show its expansibility to open-world predicate scene graph generation. Since current SGG models cannot solve this challenging task, we verify the performance of CaCao and Epic by comparing them with the naive backbone and present the comparison results fully in Table \ref{zs-result}.

Empirically, CaCao can bring out more informative predicates for better generalization. 
With the help of diverse predicates from CaCao, our Epic obtained a significant improvement of 9.6\% on novel R@100 for PredCls, verifying its effectiveness for challengeable open-world predicate SGG.
The CaCao and Epic not only improve the novel categories but also greatly improve the base categories on PredCls~(28.3\% / 31.1\% v.s. 17.6\% / 21.1\%), indicating that the entangled cross-modal prompt can provide general benefit to the representation learning of predicates, rather than merely an additional hint to the unseen predicates.


Surprisingly, even only using rich predicates generated by CaCao, Epic still performs better than VG alone. It indicates that predicates generated by CaCao are diverse and contain richer information to help SGG models learn predicate semantics for predicate generalization.
\begin{figure}[ht]
    \centering
     \vspace{-0.1cm}
   \includegraphics[width=1.0\linewidth]{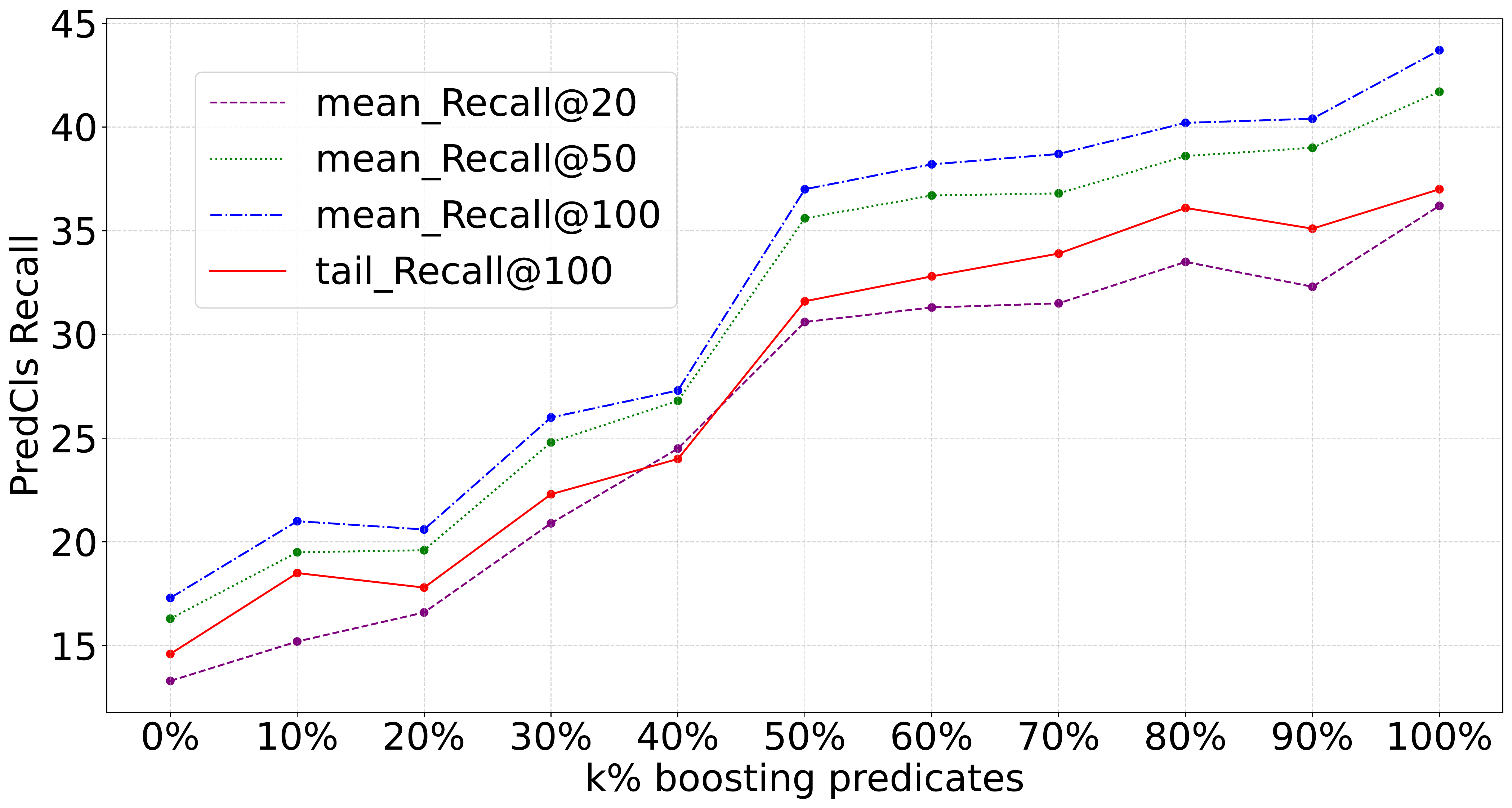}
   \caption{The influence of different proportions k\% of boosted predicates for different mR@K and tail-R@100~(\textbf{red line}).}
      \vspace{-0.4cm}
   \label{different proportions}
\end{figure}

\begin{table}[ht]
\vspace{-0.2cm}
\setlength\tabcolsep{11pt}
\resizebox{0.48\textwidth}{!}{
\begin{tabular}{clccccc}
\hline
\toprule
&\multicolumn{1}{c}{\multirow{2}{*}{Methods}} & 
\multicolumn{4}{c}{Predicate Prediction Accuracy}  \\ 
&  & ASCL& TPT& VPT & A@1/10 $\uparrow$ \\ \hline
 1 &Backbone & \XSolidBrush  & \XSolidBrush & \XSolidBrush &   0.08 / 0.21 \\ \hline
2&\quad w/o ASCL & \XSolidBrush & \Checkmark & \Checkmark &  0.38 / 0.74 \\ 
3&\quad w/o TPT  & \Checkmark & \XSolidBrush & \Checkmark &  0.47 / 0.80  \\
4&\quad w/o VPT & \Checkmark & \Checkmark & \XSolidBrush & 0.25 / 0.68  \\
\hline
8&\textbf{CaCao}  & \Checkmark & \Checkmark & \Checkmark &  \textbf{0.74 / 0.92}  \\ \hline
\end{tabular}}
\vspace{0.3mm}
\caption{Ablation study on each module of our proposed CaCao with predicate labels prediction accuracy~(\textbf{A@1/10}) metrics.}
\label{ablation}
\vspace{-0.3cm}
\end{table}
\subsection{In-depth Analysis}

\noindent\textbf{Visually-Prompted Language model.} To deeply investigate our CaCao, we further study the ablation variants of different modules in Table~\ref{ablation}. Specifically, we train the following ablation models. 1) w/o ASCL: we remove the Adaptive Semantic Cluster Loss~(ASCL). 2) w/o TPT: we remove the Textual Prompt~(TPT) in prompt templates. 3) w/o VPT: we remove the visually-prefixed Prompt~(VPT). We use Acc@1/10 as metric in Table~\ref{ablation} because it assesses the prediction accuracy of predicates from CaCao equally.

The results of Row 2 indicate that adaptive semantic cluster learning is crucial for diverse fine-grained predicate prediction. Also, the results of Row 3 validate the importance of learnable prompts on textual semantic understanding. Furthermore, Row 4 suggests that the main performance gain comes from these visual semantics contained in images~(0.25 / 0.68 $\rightarrow$ 0.74 / 0.92).


\noindent\textbf{Influence of k\% Boosting Predicates.} As shown in Figure~\ref{different proportions}, with the increase of k\% boosting predicates, the mean recall and the tail recall gradually increased in the form of overall growth. The phenomenon indicates that predicates enhanced by CaCao are all informative and consistently bring enhancements to the existing SGG models. 

\noindent\textbf{Adaptive Semantic Cluster Learning.}
Since the quality of clustering is critical for the adaptive prompt tuning in CaCao, we further explore the effect of predicate clustering under different similarity threshold initialization on the fine-grained predicates generation in Table~\ref{k_similarity_main}. Here we use A@1 as the ablation metric to clearly show the performance of predicates generation. Our observations reveal that excessively low or high similarity thresholds can lead to a decrease in predicate prediction accuracy. The possible reason is that too low similarity aggregate nearly all predicates into the same cluster and too high similarity regards each predicate individually may lead to incorrect clusters. Thus, we set the appropriate threshold as 0.7 for ASCL and obtain the optimal performance of 0.74 A@1 in CaCao.
\begin{table}[ht]
\renewcommand\arraystretch{1.5}
\resizebox{0.475\textwidth}{!}{
\begin{tabular}{c|cccccc}
\hline
 Similarity threshold & w/o ASCL & 0.1 & 0.3 & 0.5 & 0.7 & 0.9  \\ \hline
 A@1 & 0.38 & 0.39 & 0.57 & 0.63 & \textbf{0.74} & 0.48 \\ \hline
\end{tabular}
}
\vspace{0.3mm}
\caption{The influence of different predicate similarity threshold for cross-modal prompt tuning in CaCao.}
\label{k_similarity_main}
\vspace{-0.4cm}
\end{table}

\noindent\textbf{Entangled Cross-Modal Prompts.} We explore the effectiveness of the text-aware prompt and the vision-aware prompt in Epic, shown in the last two lines of Table~\ref{zs-result}. We gradually removed these entangled prompts and observed a significant decrease in performance for both base and novel classes without either prompt from another modality. These findings suggest mutual hints between the two modalities are necessary to extract associated linguistic semantics and image features for open-world predicate learning.

\noindent\textbf{Human Evaluation.} A key element of effective SGG boosting is to obtain high-quality data. Thus, we conduct a human evaluation for automatic labels from CaCao and find the radio of reasonable fine-grained predicates is \textbf{73\%}. Please refer to Appendix D for more details.

\noindent\textbf{Visualization Results.}
In Figure~\ref{visualization}, we visualize the enhancement SGG benefits from CaCao compared with the base scene graph and further present open-world predicate SGG visualization results by CaCao+Epic, intuitively illustrating the effectiveness of our proposed CaCao and Epic. The examples~(blue labels) in Figure~\ref{visualization} clearly show that the Transformer+CaCao successfully generates more fine-grained predicates than the Transformer, such as ``car-parked on-street" instead of ``car-on-street". In addition, we find that with the help of CaCao and Epic, our model can predict additional predicates~(orange labels) and even predicates of unseen categories~(red labels), such as ``building-between-street" and ``man-walk across-street".
\section{Conclusions}
In this work, we propose an automatic boosting framework CaCao that exploits linguistic knowledge from pre-trained language models to enrich existing datasets in a low-resource way.
We tackle the long-tail issue of SGG with the help of abundant informative predicates from CaCao and generalize to open-world predicate learning with the entangled cross-modal prompt design based on VL models.  
Our extensive experiments on three datasets illustrate the significant improvement of our CaCao on fine-grained scene graph generation and open-world generalization capability. 

\noindent\textbf{Acknowledgment.} This work has been supported in part by the Zhejiang NSF~(LR21F020004), the National Key R\&D Program of China~(2022ZD0160101), the NSFC~(No. 62272411),  Alibaba-Zhejiang University Joint Research Institute of Frontier Technologies, and Ant Group. We thank all the reviewers for their valuable comments.



{\small
\bibliographystyle{ieee_fullname}
\bibliography{main}
}
\clearpage

\appendix
\section{Overview}
In this supplementary material, we present:

\begin{itemize}
	\item Detailed dataset statistics in experiments~(Section \ref{1}).
	
	\item More detailed analysis of CaCao~(Section \ref{2}).
	
	
	\item Human evaluation of CaCao~(Section \ref{4}).
 
	\item Additional experimental results~(Section \ref{6}).
	
	\item Additional examples~(Section \ref{7}).

\end{itemize}

\section{Dataset Statistics}\label{1}
\noindent\textbf{Visual Genome.} Table~\ref{t1-1} and \ref{t1-3} show the coarse-grained predicates and fine-grained predicates with the number of training instances for each predicate in the Visual Genome dataset\cite{krishna2017visual}. Table~\ref{t1-2} and \ref{t1-4} show the coarse-grained predicates and fine-grained predicates with the number of training instances for each predicate after cross-modal boosting by our CaCao. We can observe that CaCao increases dataset scale, especially the tail predicates, which significantly alleviates the long-tail distribution problem in SGG.  

\noindent\textbf{GQA.} For large-scale benchmark SGG, GQA~\cite{hudson2019gqa} contains 113K images and over 3.8M relation annotations. In order to ensure the quality of the dataset, we perform a manual cleaning process to remove annotations that had poor quality or ambiguous meanings following prior works~\cite{dong2022stacked}. We finally select the top 200 object classes and top 100 predicate classes as the GQA-200 split like VG-50 to explore the generalization ability of CaCao in large-scale SGG. 

\noindent\textbf{VG-1800.} VG-1800~\cite{zhang2022fine} is another large-scale benchmark dataset, which filters out spelling errors and unreasonable relations, ultimately preserving 70,098 object classes and 1,807 predicate classes for more challenging scenarios. For each predicate category in VG-1800, there exist over 5 samples on the test set to provide a reliable evaluation.

	\begin{table*}[ht]

		\centering
		\resizebox{1.0\textwidth}{!}{
			\begin{tabular}{|c|c|c|c|c|c|c|c|c|c|c|}
				\hline
				\textbf{Coarse-grained Predicates}&above& across& against& along& and& at& behind& between& for& from\\	\hline	
				\textbf{Number of Predicates}&47341& 1996& 3092& 3624& 3477& 9903& 41356& 3411& 9145& 2945 \\ \hline
                \textbf{Coarse-grained Predicates}& has& in& in front of& near& of& on& over& to& under& with \\ \hline
				\textbf{Number of Predicates}& 277936& 251756& 13715& 96589& 146339& 712409& 9317& 2517& 22596& 66425 \\ 
				\hline 
			\end{tabular}
		}
            \vspace{0.5mm}
		\caption{Statistics of \textbf{coarse-grained predicates} for the VG-50.}
		\label{t1-1}
	\end{table*}
	\begin{table*}[!htb]

		\centering
		\resizebox{1.0\textwidth}{!}{
			\begin{tabular}{|c|c|c|c|c|c|c|c|c|c|c|}
				\hline
				\textbf{Coarse-grained Predicates}&above& across& against& along& and& at& behind& between& for& from\\	\hline	
				\textbf{Number of Predicates}&47829& 60320& 88810& 3722& 10254& 38305& 43345& 94138& 10643& 17149\\ \hline
                \textbf{Coarse-grained Predicates}& has& in& in front of& near& of& on& over& to& under& with \\ \hline
				\textbf{Number of Predicates}& 300695& 296474& 24950& 141494& 197294& 787048& 12820& 8672& 43535& 93078 \\ 
				\hline 
                
			\end{tabular}
		}
            \vspace{0.5mm}
		\caption{Statistics of \textbf{coarse-grained predicates} for the boosted VG-50 from CaCao.}
		\label{t1-2}
	\end{table*}

 \begin{table*}[!htb]

		\centering
		\resizebox{1.0\textwidth}{!}{
			\begin{tabular}{|c|c|c|c|c|c|c|c|c|c|c|}
				\hline
				\textbf{Fine-grained Predicates}&attached to& belonging to& carrying& covered in& covering& eating& flying in& growing on& hanging from& holding\\	\hline	
				\textbf{Number of Predicates}&10190& 3288& 5213& 2312& 3806& 4688& 1973& 1853& 9894& 42722\\ \hline
                \textbf{Fine-grained Predicates}& laying on& looking at& lying on& made of& mounted on& on back of& painted on& parked on& part of& playing \\ \hline
				\textbf{Number of Predicates}& 3739& 3083& 1869& 2380& 2253& 1914& 3095& 2721& 2065& 3810 \\ \hline
                \textbf{Fine-grained Predicates}& riding& says& sitting on& standing on& using& walking in& walking on& watching& wearing& wears \\ \hline
				\textbf{Number of Predicates}& 8856& 2241& 18643& 14185& 1925& 1740& 4613& 3490& 136099& 15457 \\
				\hline 
			\end{tabular}
		}
            \vspace{0.5mm}
		\caption{Statistics of \textbf{fine-grained predicates} for the VG-50.}
		\label{t1-3}
	\end{table*}
 \begin{table*}[!htb]

		\centering
		\resizebox{1.0\textwidth}{!}{
			\begin{tabular}{|c|c|c|c|c|c|c|c|c|c|c|}
				\hline
				\textbf{Fine-grained Predicates}&attached to& belonging to& carrying& covered in& covering& eating& flying in& growing on& hanging from& holding\\	\hline	
				\textbf{Number of Predicates}&80066& 20858& 79148& 54015& 17879& 100241& 6752& 20290& 90025& 68378\\ \hline
                \textbf{Fine-grained Predicates}& laying on& looking at& lying on& made of& mounted on& on back of& painted on& parked on& part of& playing \\ \hline
				\textbf{Number of Predicates}& 31783& 150817& 21944& 27189& 62583& 20628& 36882& 68218& 14727& 20789\\ \hline
                \textbf{Fine-grained Predicates}& riding& says& sitting on& standing on& using& walking in& walking on& watching& wearing& wears \\ \hline
				\textbf{Number of Predicates}& 62625& 22273& 68474& 70311& 63777& 32956& 38853& 235425& 258332& 60328 \\
				\hline 
			\end{tabular}
		}
            \vspace{0.5mm}
		\caption{Statistics of \textbf{fine-grained predicates} for the VG-50.}
		\label{t1-4}
	\end{table*}
\section{Cross-Modal Predicate Boosting}\label{2}
\begin{table}[ht]
  \centering
  \resizebox{0.475\textwidth}{!}{
  \begin{tabular}{|c|c|}
    \hline
    Predicted Predicates & Semantic Co-reference Predicates \\ \hline
`wearing'& [`wearing', `worn on', `carrying'] \\\hline 
`holding'& [`holding', `carrying', `pulling'] \\\hline
`next to'& [`next to', `sitting next to', `standing next to'] \\\hline
`standing in'& [`standing in', `standing on', `standing by']\\\hline
`below'& [`below', `beneath', `standing behind']\\\hline
`flying in'& [`flying', `flying in', `floating in']\\\hline
`sitting on'& [`sitting at', `sitting in', `is seated on']\\\hline
`hang on'& [`hang on', `hanging on', `hanging from']\\\hline
`covered in'& [`covered in', `covered with', `covered by']\\\hline
`surrounded by'& [`surrounded by', `covered by', `pulled by'] \\\hline
`walks through'&[`walks through',` is passing through', `passed by']\\\hline
  \end{tabular}
  }
  \vspace{0.5mm}
    \caption{The examples of top clustering results for semantic co-reference predicates}
    \label{coreference}
\end{table}
\begin{table}[ht]
\renewcommand\arraystretch{1.5}
		\centering
  \resizebox{0.475\textwidth}{!}{
		\begin{tabular}{l|ccc}
			\hline
\multicolumn{1}{c}{\multirow{2}*{Models}}&\textsl{PredCls}&\textsl{SGCls}&\textsl{SGDet} \\
      &zsR@50/100 $\uparrow$ &zsR@50/100 $\uparrow$ & zsR@50/100 $\uparrow$  \\
			\hline
    MOTIFS~\cite{zellers2018neural}& 10.9 / 14.5 & 2.2 / 3.0 &  0.1 / 0.2 \\
  +Resample~\cite{burnaev2015influence} & 11.1 / 14.3 &  2.3 / 3.1 & 0.1 / 0.3\\
  +TDE-GATE~\cite{tang2020unbiased} & 5.9 / 8.1 & 3.0 / 3.7 & 2.2 / 2.8\\
  +Label Refine~\cite{goel2022not} & \textbf{14.4} / \quad - \quad  & 3.0 /\quad - \quad & 3.1 /\quad - \quad \\
  +QuatRE~\cite{wang2023quaternion} &11.9 / \textbf{15.2}&2.8 / 3.6 &0.2 / 0.4\\ 
  \textbf{+CaCao}&12.0 / 13.1&\textbf{5.1 / 5.8}&\textbf{3.6 / 3.9}\\ \hline
VCTree~\cite{tang2019learning}&10.8 / 14.3 &1.9 / 2.6 &0.2 / 0.7 \\
        +TDE-GATE~\cite{tang2020unbiased} & 7.7 / 11.0 & 1.9 / 2.6 & 1.9 / 2.5\\
        +Label Refine~\cite{goel2022not} & 13.5 / \quad - \quad  &6.2 /\quad - \quad & \textbf{3.3} /\quad - \quad\\
        +QuatRE~\cite{wang2023quaternion} & 11.3 / 14.4 & 3.5 / 4.4 & 0.5 / 0.9 \\ 
        \textbf{+CaCao}&\textbf{13.6 / 14.9}&\textbf{6.5 / 7.2}&3.3 / \textbf{5.2}\\ \hline
        Transformer~\cite{tang2020unbiased}&11.3 / 14.7 & 2.5 / 3.3 & 0.9 / 1.1 \\
        \textbf{+CaCao}&\textbf{14.5 / 15.9}&\textbf{4.8 / 5.7}&\textbf{4.4 / 5.7}\\ 
			\hline
		\end{tabular}	
  }
  \vspace{0.5mm}
   \caption{Comparisons of the VG-50 SGG results on zero-shot combinational generalization performance~(zsR@K) among various approaches.}
	\label{zero-shot results}
\end{table}
\subsection{Data Preprocessing} We first collect as many detailed pictures as possible from the Internet~(\textit{i.e.} CC3M, COCO caption) as the original data for training and get nearly 80k images and 2k predicate categories with corresponding descriptions. Then we conduct semantic analysis of the corresponding description statement of each image through \textbf{Stanford CoreNLP} and preserve those informative chunks~(\textit{i.e.} V, P, N, NP, and VP) to extract fine-grained triplets contained in captions. and Since the raw data contains much noise, we further design heuristic rules~(\textit{i.e.} corpus co-occurrence frequency, layer depth of lexical analysis) to filter out predicates that are not informative or misspelling automatically instead of handling them manually. We finally eliminate those coarse-grained predicates and preserve 585 categories of diverse predicates to obtain informative~\textless subject, predicate, object\textgreater~relationships, which nearly cover most of the common situations in the real world, as shown in Table~\ref{t2}. Since the VG dataset also contains some fine-grained predicates, there are 27 categories of informative predicates we obtained have overlap with them.

 \begin{table*}[pbt]     
\centering                
\vspace{0.15cm}      
\begin{sloppypar}
\begin{tabular}{|c|c|c|}
\hline
Image & Description & Extracted Relationships \\ \hline
\multirow{5}*{\begin{minipage}[b]{0.2\columnwidth}
		\raisebox{-0.5\height}{\includegraphics[width=0.9\linewidth]{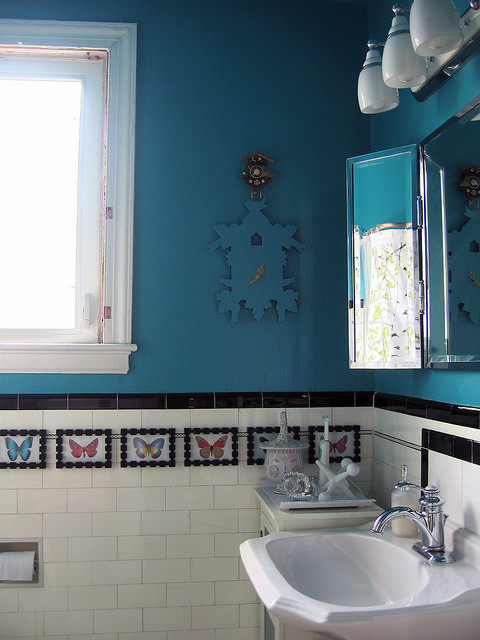}}
	\end{minipage}}  & \multirow{5}*{A clock that blends in with the wall hangs in a bathroom.} & ('clock', 'blends in with', 'wall') \\ 
                &&('clock', 'in with', 'wall') \\
                &&('clock', 'with', 'wall') \\
                &&('clock', 'hangs in', 'bathroom')\\
                &&('clock', 'in', 'bathroom')         \\ \hline               
\multirow{3}*{\begin{minipage}[b]{0.2\columnwidth}
		\raisebox{-0.5\height}{\includegraphics[width=1.1\linewidth]{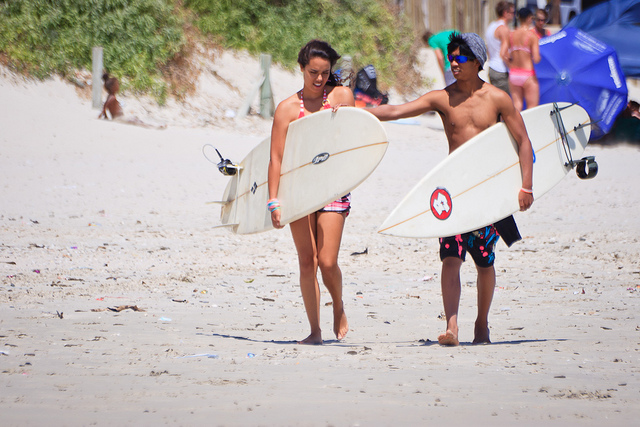}}
	\end{minipage}}  & \multirow{3}*{A couple at the beach walking with their surfboards.} & ('couple', 'at', 'beach') \\ 
                &&('couple', 'walking with', 'their-surf') \\
                &&('couple', 'with', 'their-surf')      \\ \hline 
\multirow{4}*{\begin{minipage}[b]{0.2\columnwidth}
		\raisebox{-0.5\height}{\includegraphics[width=0.7\linewidth]{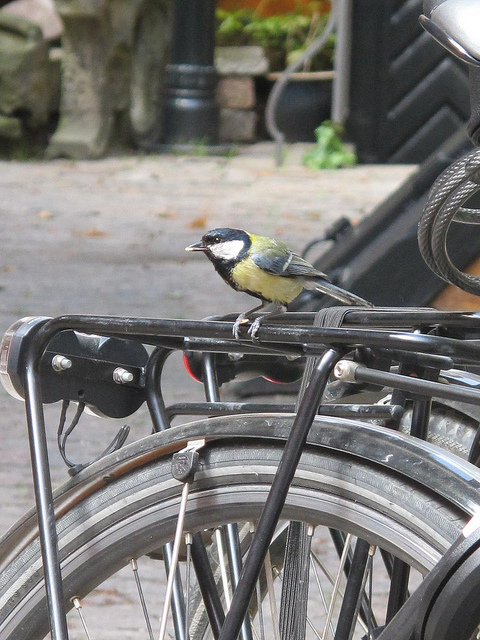}}
	\end{minipage}}  & \multirow{4}*{A yellow and black bird standing on and hanging with a bike rack.} & ('black-bird', 'on', 'bike-rack') \\
                    && ('yellow-bird', 'on', 'bike-rack') \\
                &&('black-bird', 'standing on', 'bike-rack') \\
                &&('black-bird', 'hanging with', 'bike-rack')      \\ \hline 
                
\end{tabular}
\end{sloppypar}
\vspace{0.8mm}
\caption{The examples of ~\textless subject, predicate, object\textgreater~ extraction from raw data for prompt tuning.}
\label{t2}
\end{table*}

\begin{figure*}[ht]
\begin{subfigure}{1.0\linewidth}
    \centering
    \includegraphics[width=1.\linewidth]{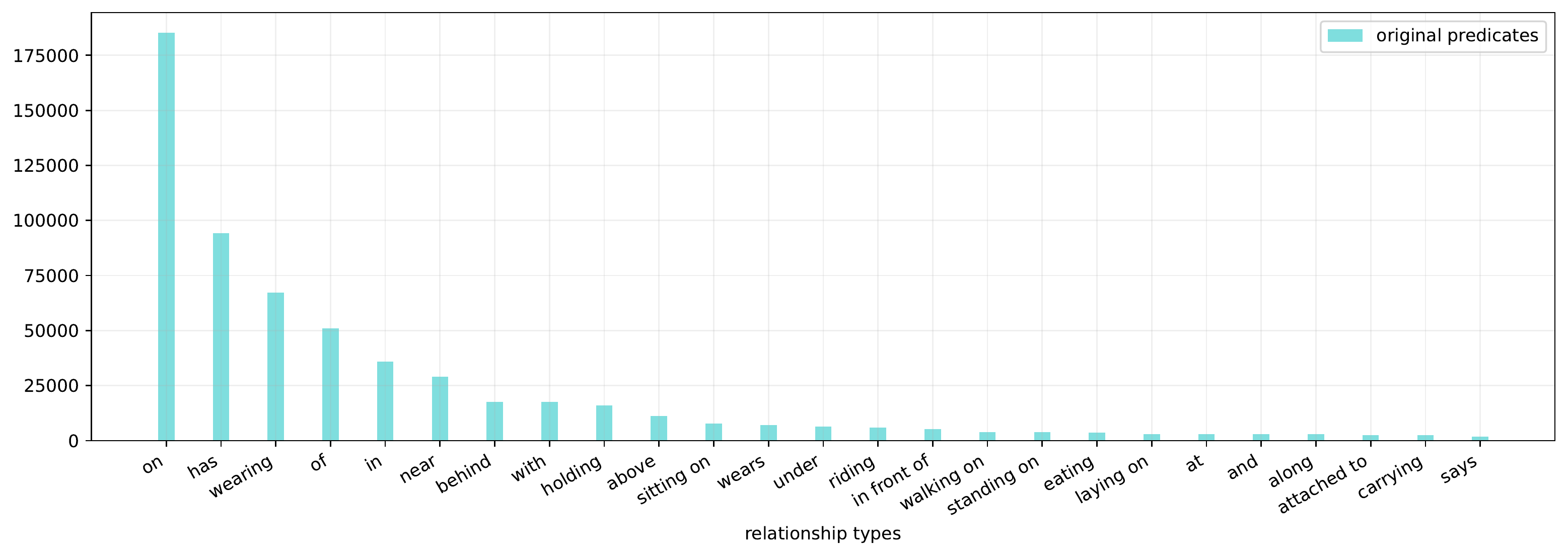}
    \caption{Top-25 predicates of original distribution}
    \label{original-distribution}
  \end{subfigure}
\begin{subfigure}{1.0\linewidth}
    \centering
    \includegraphics[width=1.0\linewidth]{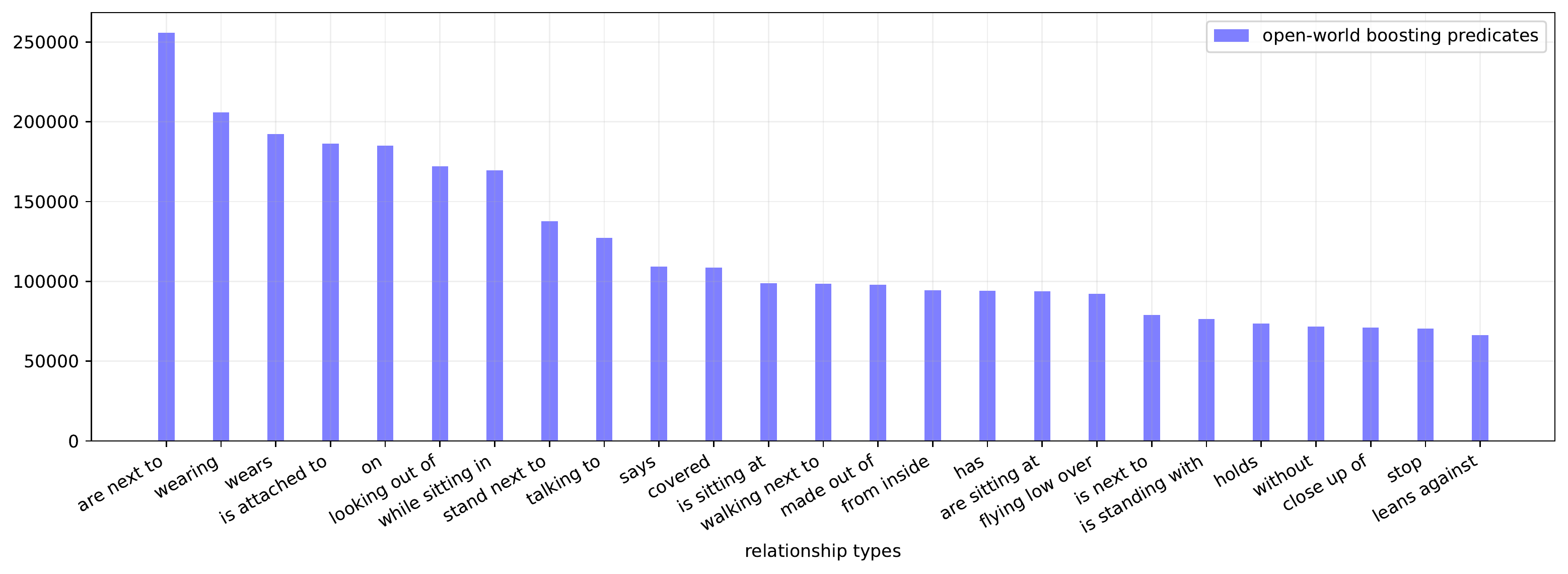}
    \caption{Top-25 predicates of open-world distribution}
    \label{open-distribution}
     \end{subfigure}
     \caption{Qualitative predicate distributions of the standard SGG dataset and the open-world enhanced data from CaCao.}
\end{figure*}


\subsection{Adaptive Semantic Cluster Loss}
\noindent\textbf{Importance of semantic co-reference.} We list more semantic co-reference words and some clustering results as shown in the table~\ref{coreference}, such as he ``walks through" / ``is passing through" / ``passed by" a street may correspond to the same predicate ``walking on. 
To address the semantic co-reference challenge, we proceed to train CaCao using the ASCL based on predicate semantic clusters. Since there are strong dependencies between triples in complex scenarios, for each predicate class, we represent and average the embeddings of all triples corresponding to it. To achieve this, we use the feature map of the last BERT layer as the representation of each entire triplet. We initialize the target predicate according to different similarity thresholds, and then confirm the number of initial centroids.

\noindent\textbf{Importance of semantic ambiguity.} Although semantic clustering is static to contexts, CaCao dynamically adjusts the predicted results based on context-aware labels, which are sensitive to various contexts. Then semantic clustering promotes diverse expressions for the adjusted synonyms, which are also context-sensitive. Besides, we find only a few semantic ambiguities caused by contexts~(6\% for `wearing' to `has') in the current dataset and analyze that the influence of contexts on synonyms in SGG is small during training. For a few failure cases caused by complex semantic ambiguities, we provide several candidates to correct the mapping and obtain more accurate prediction results.

\subsection{Fine-Grained Predicate Boosting}
In Figure~\ref{original-distribution} and~\ref{open-distribution}, we show the predicate distributions of the standard SGG dataset and open-world boosted data from CaCao. To enhance predicates into the target scene graphs, we need to establish the mapping from diversity predicates to target predicates, as shown in Table~\ref{t5}.

Moreover, we notice that there exists ambiguity and overlap between coarse-grained predicates and fine-grained predicates in fact. We further create the mapping between fine-grained predicates and coarse-grained predicates based on the semantic association between predicates~\cite{chen2019scene}. We then figure out those low-confidence fine-grained predicates and map them into general predicates as final predicted results to achieve better trade-offs on long-tail recognition.


\section{Human Evaluation}\label{4}
A key element of effective SGG boosting is to obtain high-quality data. Thus, we conduct a human evaluation for automatically obtained labels from CaCao to verify the quality. We randomly select 100 images containing 545 base relationships and 3543 novel relationships to validate the accuracy and  informativeness of the predicates associated with these augmented relationships, ensuring their utility in facilitating open-world predicate scene graph generation. We show the result in Table~\ref{t6}. We observe the radio of reasonable fine-grained predicates in CaCao is \textbf{73.4\%} and the proportion of coarse-grained predicates is greatly reduced by CaCao's enhanced predicates. Consequently, the results indicate that the predicates enhanced by CaCao can effectively provide fine-grained information.


\begin{table*}[!htb]
\renewcommand\arraystretch{1.5}
	\setlength\tabcolsep{10pt}

		\centering
		\resizebox{1.0\textwidth}{!}{
			\begin{tabular}{ccccc}
				\hline
				&\textbf{Total Predicate} & \textbf{True Predicate} & \textbf{Fine-Grained Predicate~(\%)} $\uparrow$ & \textbf{Coarse-Grained Predicate~(\%)} $\downarrow$ \\	\hline
            Original & 545 & 545 & 119~(21.8\%) & 426~(78.2\%) \\ 
            \textbf{CaCao} & 3543 & 2427 & \textbf{1781~(73.4\%)} & \textbf{646~(26.6\%)}\\ \hline
            Overall & 4088 & 2972& 1900~(63.9\%) &1072~(36.1\%)\\
				\hline 
                
			\end{tabular}
		}
    \vspace{0.5mm}
    \caption{Human evaluation for the accuracy and variety of enhanced predicates from CaCao.}
    \label{t6}
  \end{table*}
\begin{table*}[ht]
\setlength\tabcolsep{10pt}
\resizebox{1.0\textwidth}{!}{
\begin{tabular}{cccccc}
\toprule
&\multicolumn{1}{c}{\multirow{2}{*}{Model Type}} & \multirow{2}{*}{Methods}  & \multicolumn{3}{c}{Scene Graph Detection} \\ 
 & & & R@50/100 $\uparrow$ & mR@50/100 $\uparrow$ & F@50/100 $\uparrow$ \\ 
 \hline
&\multicolumn{1}{c}{\multirow{3}{*}{Specific}} & \multicolumn{1}{l}{BGNN~\cite{li2021bipartite}}  & 31.0 / 35.8& 10.7 / 12.6 &15.9 / 18.6 \\
&& \multicolumn{1}{l}{SVRP~\cite{he2022towards}} &  31.8 / 35.8 & 10.5 / 12.8 & 15.8 / 18.9 \\
&& \multicolumn{1}{l}{DT2-ACBS~\cite{desai2021learning}} &  15.0 / 16.3 & 22.0 / 24.0 & 17.8 / 19.4 \\
\hline\hline

&\multicolumn{1}{c}{\multirow{2}{*}{One-stage}} & \multicolumn{1}{l}{SSRCNN~\cite{teng2022structured}} & 23.7 / 27.3 &  18.6 / 22.5 &  20.8 / 24.7\\
&&\multicolumn{1}{l}{\quad\textbf{+CaCao~(ours)}} & 25.4 / 30.0 & \textbf{18.7 / 23.1} & \textbf{21.5 / 26.1}\\
\hline\hline

\multirow{10}{*}{\rotatebox[]{90}{Model-Agnostic strategy}}& & \multicolumn{1}{l}{Motif~\cite{zellers2018neural}} &   31.0 / 35.1 & 6.7 / 7.7 & 11.0 / 12.6   \\
&\multicolumn{1}{c}{Resample} & \multicolumn{1}{l}{\quad+Resample~\cite{burnaev2015influence}} &   30.5 / 35.4  &  8.2 / 9.7 & 12.9 / 15.2 \\ 
&\multicolumn{1}{c}{\multirow{4}{*}{Reweight}}& \multicolumn{1}{l}{\quad+Reweight~\cite{wang2021seesaw}} &  24.4 / 29.3 & 10.5 / 13.2 & 14.7 / 18.2\\
&& \multicolumn{1}{l}{\quad+CogTree~\cite{yu2020cogtree}} & 20.0 / 22.1 & 10.4 / 11.8 & 13.7 / 15.4\\
&&\multicolumn{1}{l}{\quad+FGPL~\cite{lyu2022fine}} &21.3 / 24.3 & 15.4 / 18.2 & 17.9 / 20.8\\
&&\multicolumn{1}{l}{\quad+GCL~\cite{dong2022stacked}} & 18.4 / 22.0 &16.8 / 19.3& 17.6 / 20.6\\
&\multicolumn{1}{c}{Causal Rule}& \multicolumn{1}{l}{\quad+TDE~\cite{tang2020unbiased}} &  16.9 / 20.3 & 8.2 / 9.8 & 11.0 / 13.2\\ \cline{2-6}
&\multicolumn{1}{c}{\multirow{4}{*}{Data Enhancement}}& \multicolumn{1}{l}{\quad+Only Caption Relations} & 20.3 / 25.0 & 8.2 / 10.0 & 11.7 / 14.3\\
&&\multicolumn{1}{l}{\quad+DLFE~\cite{chiou2021recovering}} &  25.4 / 29.4 & 11.7 / 13.8 & 16.0 / 18.8\\
&&\multicolumn{1}{l}{\quad+IETrans~\cite{zhang2022fine}} &\cellcolor{lightgray}23.5 / 27.2 & \cellcolor{lightgray}15.5 / 18.0 & \cellcolor{lightgray}18.7 / 21.7 \\ 
&&\multicolumn{1}{l}{\quad\textbf{+CaCao~(ours)}} & \cellcolor{lightgray}\textbf{24.4 / 29.1} & \cellcolor{lightgray}\textbf{17.1 / 20.0} & \cellcolor{lightgray}\textbf{20.5 / 23.7} \\
\bottomrule
\end{tabular}
}
\vspace{3mm}
\caption{Performance~(\%) of our method \textbf{CaCao} and other baselines with different model types for both \textbf{head} and \textbf{tail} categories on VG-50 dataset.}
\label{head-result}
\end{table*}

\section{Additional Experiment Analyses}\label{6}
\noindent\textbf{Compositional Generalization.} Thanks to the remarkable performance of our CaCao in the open-world scenario, it demonstrates the potential to improve the model compositional generalization ability in traditional zero-shot scene graph generation tasks~\cite{lu2016visual, goel2022not,wang2023quaternion}. Table~\ref{zero-shot results} presents the zero-shot Recall@K metrics in each task~(\textsl{i.e.}, \textit{PredCls}, \textit{SGCls}, and \textit{SGDet}), providing a comprehensive evaluation of the compositional generalization performance. We compare our proposed CaCao with other state-of-the-art approaches. Our proposed method achieves improvements in most of the settings with different SGG backbones, except for MOTIFS in PredCls. MOTIFS being a textual-only model fails to effectively utilize the enhanced data to learn implicit features for discerning the combination of relations and hence performs poorly when given the ground truth contexts. Conversely, the multi-modal VCTree and Transformer models effectively utilize extra triplet-level data due to their ability to align more visual information, facilitating generalization to unseen triplets during testing.



 \noindent\textbf{Further Evaluation on Head and Tail Predicates.} Since CaCao brings much extensive visual relation knowledge on various visual predicates from powerful VL-models, the CaCao may achieve a better trade-off on long-tail distribution SGG. Our results on the whole category set partly give evidence that CaCao can achieve a better balance in the long-tail distribution. Additionally, we inspect the performance of CaCao across non-rare head predicates to further verify its better balance between head and tail predicate categories in Table \ref{head-result} \textbf{R@K}. Following prior works~\cite{zhang2022fine}, we further use the harmonic average of R@K and mR@K to jointly evaluate R@K and mR@K, which is denoted as \textbf{F@K}. From Table \ref{head-result}, we observe that \textbf{CaCao} outperforms other SOTA model-agnostic methods and specific string baseline according to the joint metric F@K~(\textbf{20.5 / 23.7} of F@50/100 on SGDet), showing the effectiveness of CaCao on both head and tail categories.

\section{Additional Examples}\label{7}
\begin{figure*}
    \begin{subfigure}{1.0\linewidth}
        
    \centering
    \includegraphics[width=1.\linewidth]{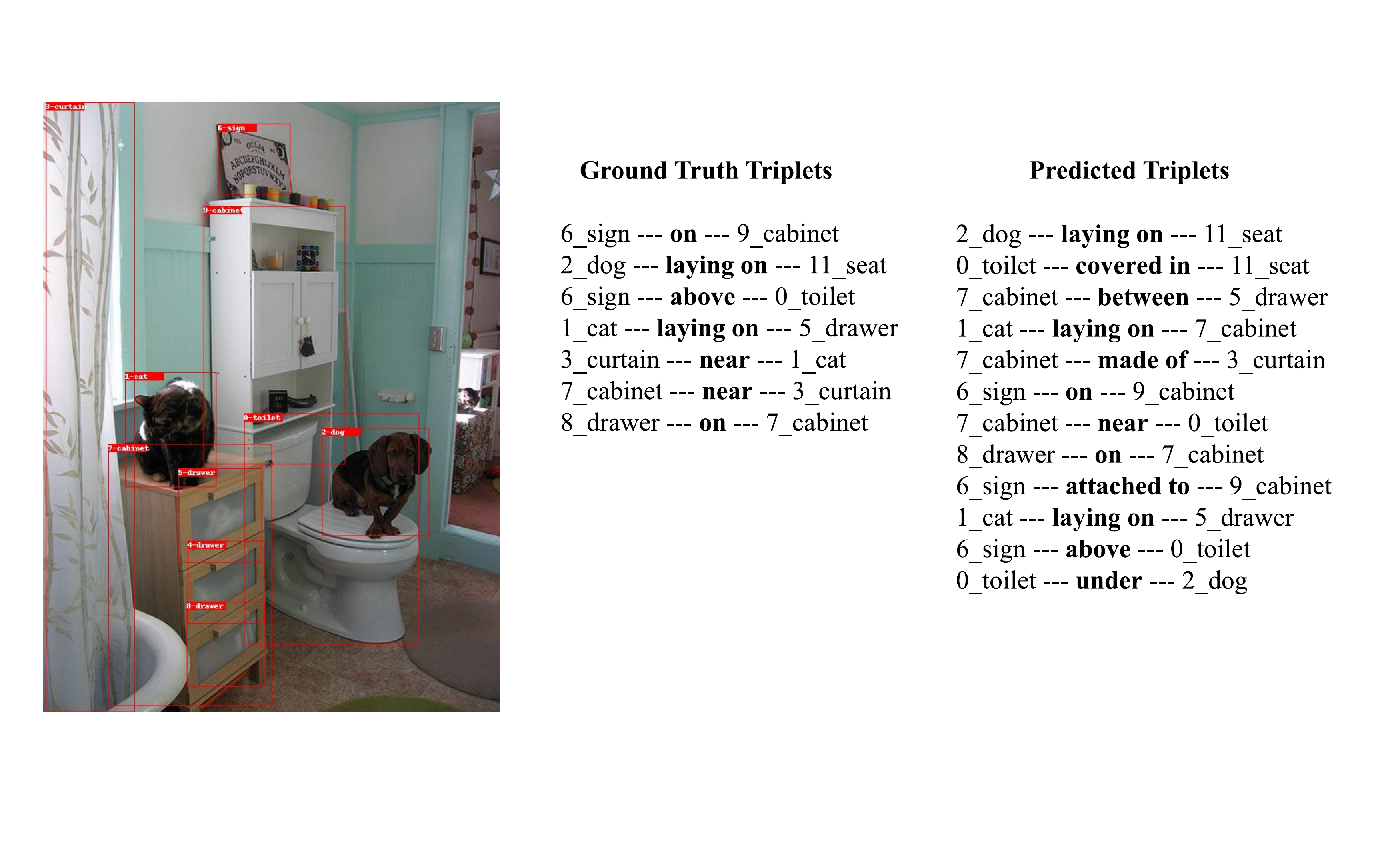}
   \end{subfigure}
    \begin{subfigure}{1.0\linewidth}
    \centering
    \includegraphics[width=1.0\linewidth]{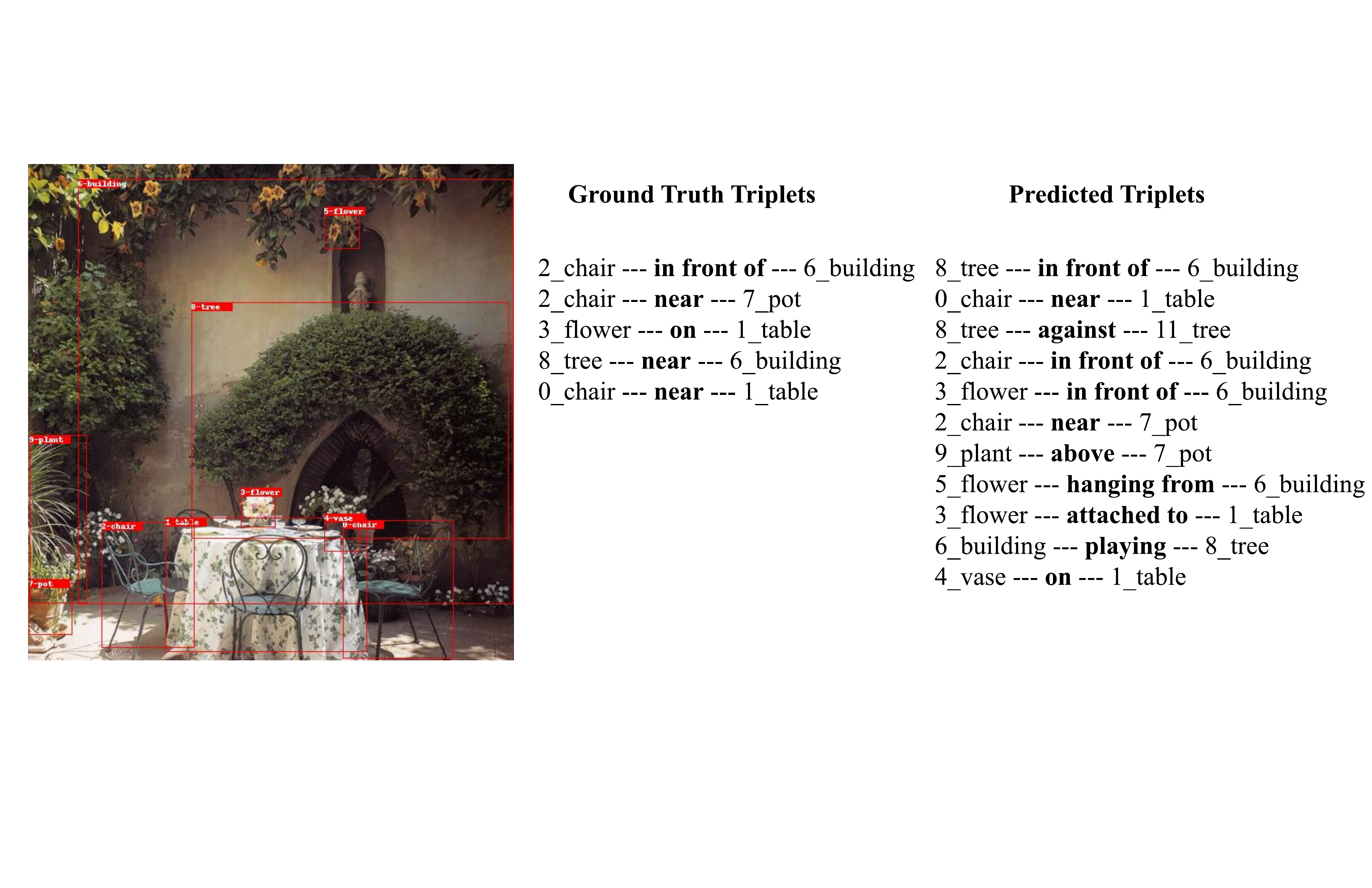}
     \end{subfigure}
     \caption{Additional Qualitative Results for Transformer equipped with our CaCao framework for predicate enhancement with the ground truth relationships. The predicted triplets are from the SGDet setting.}
     \label{f2}
\end{figure*}

Figure~\ref{f2} shows some more examples for qualitative visualizations of enhanced SGG based on our CaCao.
\clearpage
\begin{table*}
  \centering
  \begin{tabular}{|c|}
    \hline
   \textbf{Open-world predicate relationships $\rightarrow$ Target predicate relationships} \\ \hline

[`sidewalk', `\textit{in between}', `car']~$\rightarrow$~[`sidewalk', `\textit{between}', `car']\\\hline
[`sidewalk', `\textit{walking across}', `street']~$\rightarrow$~[`sidewalk', `\textit{across}', `street']\\\hline
[`tree', `\textit{hanging in}', `building']~$\rightarrow$~[`tree', `\textit{hanging from}', `building']\\\hline
[`tree', `\textit{uses}', `phone']~$\rightarrow$~[`tree', `\textit{using}', `phone']\\\hline
[`car', `\textit{are parked on}', `street']~$\rightarrow$~[`car', `\textit{parked on}', `street']\\\hline
[`street', `\textit{parked at}', `sidewalk']~$\rightarrow$~[`street', `\textit{parked on}', `sidewalk']\\\hline
[`street', `\textit{among}', `car']~$\rightarrow$~[`street', `\textit{between}', `car']\\\hline
[`phone', `\textit{hanging on}', `tree']~$\rightarrow$~[`phone', `\textit{hanging from}', `tree']\\\hline
[`motorcycle', `\textit{displaying}', `person']~$\rightarrow$~[`motorcycle', `\textit{carrying}', `person']\\\hline
[`building', `\textit{connected to}', `pole']~$\rightarrow$~[`building', `\textit{attached to}', `pole']\\\hline
[`street', `\textit{parked at}', `sidewalk']~$\rightarrow$~[`street', `\textit{parked on}', `sidewalk']\\\hline
[`shirt', `\textit{leans against}', `woman']~$\rightarrow$~[`shirt', `\textit{against}', `woman']\\\hline
[`glass', `\textit{hanging on}', `head']~$\rightarrow$~[`glass', `\textit{hanging from}', `head']\\\hline
[`chair', `\textit{to make}', `leg']~$\rightarrow$~[`chair', `\textit{made of}', `leg']\\\hline
[`man', `\textit{watch}', `woman']~$\rightarrow$~[`man', `\textit{watching}', `woman']\\\hline
[`man', `\textit{leaning up against}', `table']~$\rightarrow$~[`man', `\textit{against}', `table']\\\hline
[`screen', `\textit{laying on}', `paper']~$\rightarrow$~[`screen', `\textit{lying on}', `paper']\\\hline
[`paper', `\textit{looking up at}', `screen']~$\rightarrow$~[`paper', `\textit{looking at}', `screen']\\\hline
[`tree', `\textit{hanging over}', `trunk']~$\rightarrow$~[`tree', `\textit{hanging from}', `trunk']\\\hline
[`car', `\textit{hooked up to}', `pole']~$\rightarrow$~[`car', `\textit{attached to}', `pole']\\\hline
[`tree', `\textit{across from}', `fence']~$\rightarrow$~[`tree', `\textit{between}', `fence']\\\hline
[`sidewalk', `\textit{hanging in}', `trunk']~$\rightarrow$~[`sidewalk', `\textit{hanging from}', `trunk']\\\hline
[`sidewalk', `\textit{traveling on}', `leaf']~$\rightarrow$~[`sidewalk', `\textit{growing on}', `leaf']\\\hline
[`boy', `\textit{looking down at}', `car']~$\rightarrow$~[`boy', `\textit{looking at}', `car']\\\hline
[`woman', `\textit{is using}', `pant']~$\rightarrow$~[`woman', `\textit{using}', `pant']\\\hline
[`woman', `\textit{towing}', `shirt']~$\rightarrow$~[`woman', `\textit{carrying}', `shirt']\\\hline
[`head', `\textit{connected to}', `nose']~$\rightarrow$~[`head', `\textit{attached to}', `nose']\\\hline
[`hair', `\textit{is looking at}', `child']~$\rightarrow$~[`hair', `\textit{looking at}', `child']\\\hline
[`nose', `\textit{tied to}', `head']~$\rightarrow$~[`nose', `\textit{attached to}', `head']\\\hline
[`finger', `\textit{is parked on}', `hand']~$\rightarrow$~[`finger', `\textit{painted on}', `hand']\\\hline
[`man', `\textit{eaten}', `pizza']~$\rightarrow$~[`man', `\textit{eating}', `pizza']\\\hline
[`windshield', `\textit{towing}', `umbrella']~$\rightarrow$~[`windshield', `\textit{carrying}', `umbrella']\\\hline
[`airplane', `\textit{hanging on}', `wing']~$\rightarrow$~[`airplane', `\textit{hanging from}', `wing']\\\hline
[`airplane', `\textit{hanging on}', `wing']~$\rightarrow$~[`airplane', `\textit{hanging from}', `wing']\\\hline
[`airplane', `\textit{flying high in}', `sky']~$\rightarrow$~[`airplane', `\textit{flying in}', `sky']\\\hline
[`sign', `\textit{strapped}', `arrow']~$\rightarrow$~[`sign', `\textit{on}', `arrow']\\\hline
[`face', `\textit{connected to}', `neck']~$\rightarrow$~[`face', `\textit{above}', `neck']\\\hline
[`tree', `\textit{across from}', `building']~$\rightarrow$~[`tree', `\textit{across}', `building']\\\hline
[`roof', `\textit{across from}', `building']~$\rightarrow$~[`roof', `\textit{along}', `building']\\\hline
[`jacket', `\textit{is cluttered with}', `man']~$\rightarrow$~[`jacket', `\textit{with}', `man']\\\hline
[`sign', `\textit{are showing on}', `building']~$\rightarrow$~[`sign', `\textit{says}', `building']\\\hline
[`short', `\textit{in between}', `man']~$\rightarrow$~[`short', `\textit{with}', `man']\\\hline
[`jean', `\textit{stacked on}', `man']~$\rightarrow$~[`jean', `\textit{painted on}', `man']\\\hline
[`person', `\textit{is walking on}', `sidewalk']~$\rightarrow$~[`person', `\textit{walking on}', `sidewalk']\\\hline
[`chair', `\textit{are looking at}', `boy']~$\rightarrow$~[`chair', `\textit{in front of}', `boy']\\\hline

  \end{tabular}
  \vspace{0.5mm}
    \caption{Examples of open-world predicates to target predicates mapping.}
    \label{t5}
\end{table*}

\end{document}